%% file: main.tex
\documentclass{article}
\usepackage[preprint]{neurips_2026}
\setcitestyle{numbers,comma,sort&compress}

\usepackage[utf8]{inputenc}
\usepackage[T1]{fontenc}
\usepackage[bookmarks=false,breaklinks=true]{hyperref}
\usepackage{url}
\usepackage{booktabs}
\usepackage{amsfonts}
\usepackage{amsmath}
\usepackage{amssymb}
\usepackage{nicefrac}
\usepackage{microtype}
\usepackage{xcolor}
\usepackage{graphicx}
\usepackage{subcaption}
\usepackage{wrapfig}
\usepackage{multirow}
\usepackage{enumitem}
\usepackage{xspace}
\usepackage{longtable}
\usepackage{array}
\usepackage{tabularx}
\usepackage{etoc}
\usepackage{placeins}
\usepackage{etoolbox}
\usepackage{framed}

\newcommand{\finevlaappendixtoc}{%
  \begingroup
  \etocsettocstyle{\subsection*{Contents}}{}
  \etocsetnexttocdepth{subsubsection}
  \localtableofcontents
  \endgroup
}

\newcommand{\finevlaappendixfloatbarriers}{%
  \pretocmd{\subsection}{\FloatBarrier}{}{}%
  \pretocmd{\subsubsection}{\FloatBarrier}{}{}%
}

\input{commands}

\newcommand{\affxstar}{\textsuperscript{*\textcolor{purple}{\textbf{x}}}}
\newcommand{\affx}{\textsuperscript{\textcolor{purple}{\textbf{x}}}}
\newcommand{\affq}{\textsuperscript{\textcolor{teal}{\textbf{q}}}}

\title{FineVLA: Fine-Grained Instruction Alignment \\ for Steerable Vision-Language-Action Policies}

\author{%
\textbf{Xintong Hu}\affxstar \
\textbf{Xuhong Huang}\affxstar \
\textbf{Jinyu Zhang}\affx \
\textbf{Yutong Yao}\affx \
\textbf{Yuchong Sun}\affq
\\[2pt]
\textbf{Qiuyue Wang}\affq \
\textbf{Mingsheng Li}\affq \
\textbf{Sicheng Xie}\affq \
\textbf{Yitao Liu}\affx \
\textbf{Junhao Chen}\affx
\\[2pt]
\textbf{Yixuan Chen}\affx \
\textbf{Yingming Zheng}\affx \
\textbf{Shuai Bai}\affq \
\textbf{Tao Yu}\textsuperscript{\dag}\affx
\\[8pt]
\affx\,XLANG Lab, The University of Hong Kong \quad
\affq\,Qwen Team, Alibaba Inc.
\\[4pt]
\raisebox{-1.5pt}{\includegraphics[height=10pt]{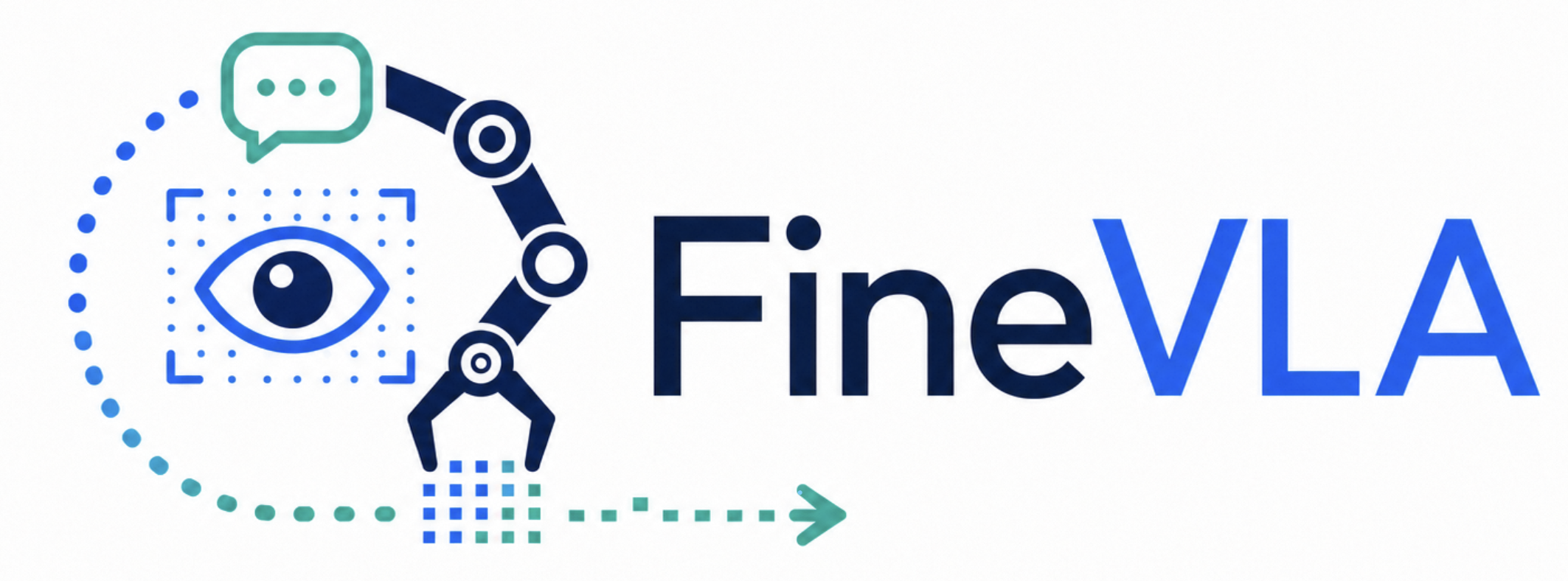}}%
\hspace{2pt}\href{https://finevla.xlang.ai/}{Project Page: https://finevla.xlang.ai/}
}

\newcommand{\authorfootnote}{%
  \footnotetext[0]{\kern-1.8em$^{*}$\,Equal contribution.\quad $^{\dagger}$\,Corresponding authors.}
}

\begin{document}
\raggedbottom

\begin{center}
\begin{minipage}[c]{0.45\textwidth}
  \includegraphics[height=20pt]{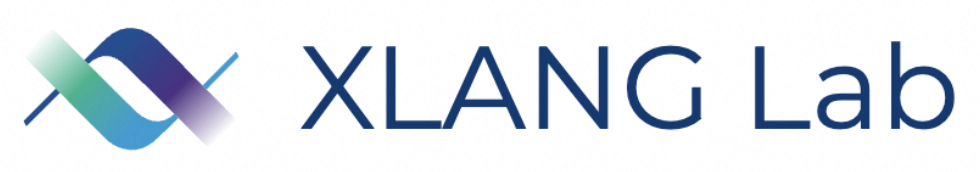}
\end{minipage}%
\hfill
\begin{minipage}[c]{0.45\textwidth}
  \raggedleft
  \includegraphics[height=28pt]{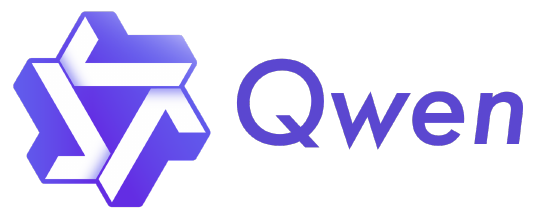}
\end{minipage}
\end{center}
\vspace{-8pt}

\maketitle
\authorfootnote

\begin{abstract}
\input{sections/abstract}
\end{abstract}

\input{sections/introduction}        
\input{sections/dataset}             
\input{sections/method}              
\input{sections/experiment}           
\input{sections/analysis}            
\input{sections/related_work}        
\input{sections/conclusion}          


\clearpage
\bibliographystyle{unsrtnat}
\bibliography{references}

\clearpage
\appendix
\finevlaappendixfloatbarriers
\section{Appendix}
\label{app:appendix}
\label{app:roadmap}
\label{sec:app_roadmap}
\finevlaappendixtoc
\input{sections/appendix}


\end{document}

%% file: commands.tex
\newcommand{\framework}{FineVLA}                    
\newcommand{\toolname}{FineVLA-Tool}                
\newcommand{\vlmname}{RoboFine-VLM}                  
\newcommand{\benchmarkname}{RoboFine-Bench}          
\newcommand{\policyname}{FineVLA-Policy}            
\newcommand{\datasetname}{FineVLA-Data}              
\newcommand{\qwenvl}{Qwen3.5-4B}



%% file: sections/abstract.tex
Vision-Language-Action (VLA) models are increasingly expected to not only
complete robot tasks, but also follow human instructions about \emph{how} those
tasks should be executed. However, existing robot datasets usually pair
trajectories with coarse goal-level language, leaving execution-critical
details such as active arm, approach direction, and contact
region unspecified. This limits steerable policy
learning and robotic video understanding.
We introduce \textbf{\framework{}}, an open framework for action-aligned
fine-grained VLA supervision. The framework includes: (1) a data construction
tool that unifies 972,247 trajectories across 85K tasks from 10 open-source
robot datasets and builds \textbf{\datasetname{}}, a human-verified dataset of
47,159 fine-grained trajectories; (2) a held-out benchmark with 500 videos,
11,631 atomic facts, and 1,030 VQA questions; (3) a robotics-specialized VLM
annotator for scalable fine-grained annotation; and
(4) a steerable \textbf{VLA policy} trained with controlled mixtures of fine-grained and
raw goal-level instructions.
Our policy experiments yield three findings. First, fine-grained supervision
does not sacrifice goal-level task success: fine-grained-only
(FG-only) improves over raw-instruction-only (Raw-only) by
+1.4 to +8.1 success-rate points across architecture and data-scale settings.
Second, fine-grained and raw instructions are complementary: performance follows
a consistent inverted-U trend, peaking around
FG\,:\,Raw\,=\,1\,:\,2 to 1\,:\,1. The strongest mixed setting reaches
\textbf{86.8\%}/\textbf{82.5\%} in RoboTwin simulation and \textbf{62.7}/100 in
real-world dual-arm manipulation, compared with 49.9 for Raw-only. Third,
fine-grained supervision directly improves steerable control
by increasing compliance with language-specified execution
factors: in real-world evaluation, the largest gains over
Raw-only appear on pose (+23), color (+18), and approach
direction (+18)---factors where goal-level instructions
provide no guidance. Overall, fine-grained language should
augment goal-level instructions: specifying \emph{how} to
execute alongside \emph{what} to achieve.

%% file: sections/introduction.tex
\section{Introduction}

Vision-Language-Action (VLA) models are moving from task-level
robot control toward policies that can be \emph{steered} by human
instructions. In this work, we use \emph{steerability} to mean the
ability to execute the same high-level goal in different ways
according to user-specified execution constraints, such as which
arm to use, which target object to manipulate, how to approach it,
where to make contact, which motion or rotation direction to
follow, and what final configuration to achieve. Recent robot
foundation models such as $\pi_{0.7}$, LingBot-VLA, GR00T N1.7,
and
GEN-1~\citep{pi0_7,lingbot_vla,isaac_gr00t_github,gen1_blog_2026}
suggest that future robot policies should not only infer
\emph{what} task to complete, but also follow instructions about
\emph{how} the task should be performed.

However, building open, steerable VLA systems remains challenging
for three reasons.
\textbf{(i)~Heterogeneous data and missing fine-grained annotation
infrastructure.}
Existing open-source robot datasets use diverse action and state
representations that cannot be directly
unified~\citep{oxe,RDT-1B}. Within the same task, demonstrations
are heavily redundant, and most trajectories carry only a single
goal-level description (e.g., ``pick up the cup'') while the
execution process---actor choice, approach direction, contact
region, motion path, and state transitions---remains unspecified.
The problem is not only that labels are coarse, but that there is
still no open infrastructure for producing action-aligned
fine-grained supervision at scale.
\textbf{(ii)~Lack of benchmarks and scalable annotators for
fine-grained robotic video understanding.}
Although general video-language models and dense captioning methods
have advanced video description, their captions often focus on
scene appearance rather than action-relevant execution details
such as contact regions, approach directions, and motion paths.
Existing embodied
benchmarks~\citep{robovqa,robobench,handyvqa} mainly evaluate
spatial reasoning or hand-object dynamics, but do not
systematically measure whether VLMs capture process-level
manipulation details. There is also a lack of open,
robotics-specialized annotators for action-aligned fine-grained
captions.
\textbf{(iii)~Unknown effectiveness and training recipe.}
Even if fine-grained data were available, the community lacks
systematic evidence on whether action-aligned instructions improve
policy learning, and what mixture of fine-grained and goal-level
supervision yields the best steerable control.

\begin{figure}[!t]
  \centering
  \includegraphics[width=\linewidth]{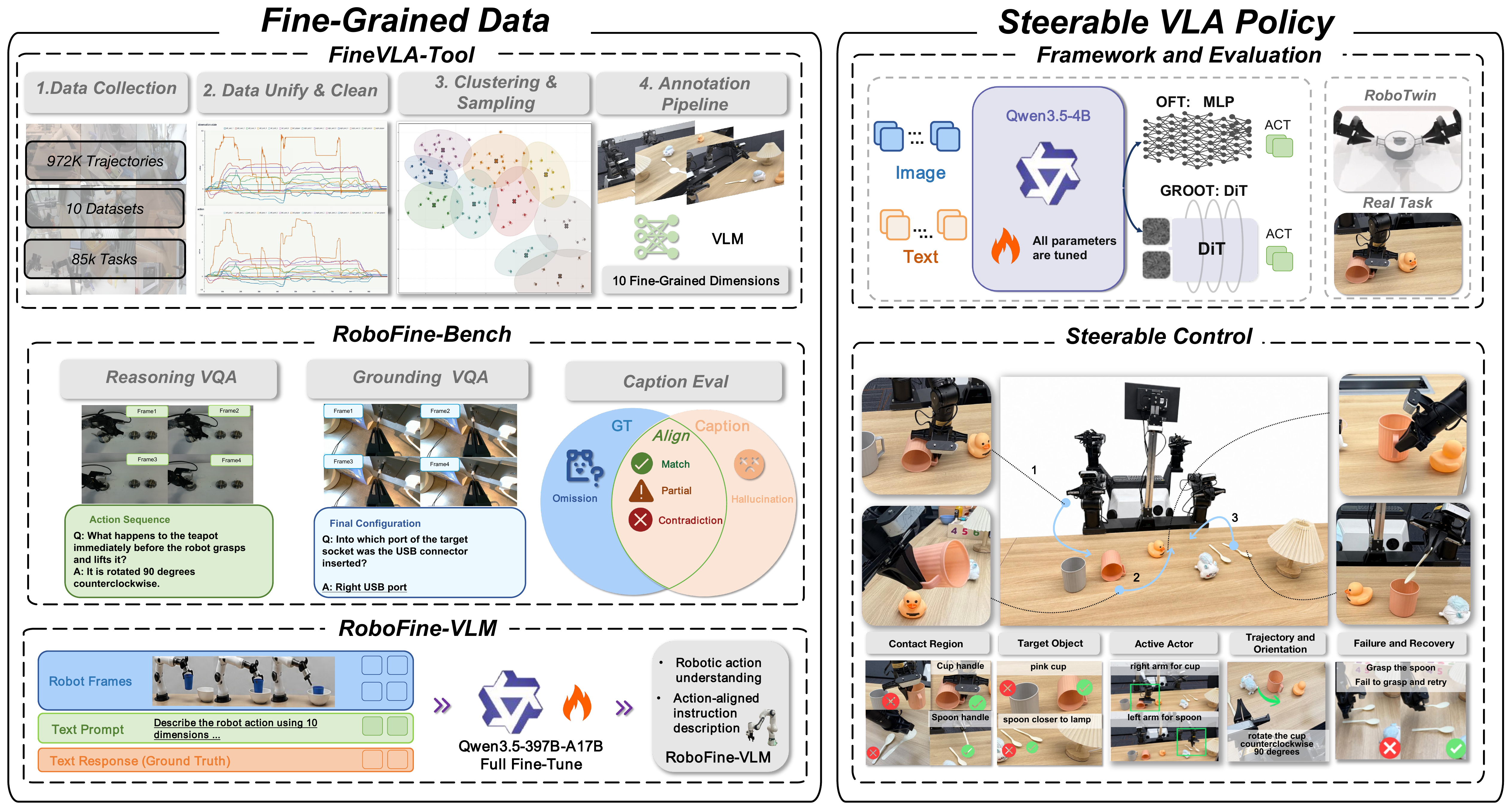}
  \caption{\textbf{Overview of \framework{}.}
  \framework{} builds a closed loop for action-instruction alignment,
  connecting fine-grained data construction, robotic video
  understanding, scalable annotation, and steerable VLA policy
  learning.
  \textbf{Left}: \toolname{} unifies heterogeneous robot trajectories
  from 10 open-source datasets, removes redundant demonstrations
  through clustering and sampling, and annotates representative
  trajectories with action-aligned descriptions across ten
  fine-grained dimensions. The resulting \datasetname{} supports
  both \benchmarkname{}, which evaluates fine-grained robotic video
  understanding through Grounding VQA, Reasoning VQA, and Caption
  Evaluation, and \vlmname{}, a robotics-specialized VLM trained as
  a scalable annotator for new trajectories.
  \textbf{Right}: \policyname{} is trained with mixtures of raw
  goal-level instructions and fine-grained process-level
  instructions under two action-decoding architectures, and is
  evaluated in both RoboTwin simulation and real-world dual-arm
  manipulation. The steerable-control examples illustrate how
  fine-grained language specifies execution-sensitive factors such
  as contact region, target object, active actor, trajectory and
  orientation, and failure recovery.}
  \label{fig:main_figure}
\end{figure}

To address these challenges, we introduce \textbf{\framework{}},
a fully open-source framework for scaling fine-grained VLA data,
robotic video understanding, and steerable VLA policies
(Figure~\ref{fig:main_figure}).
The framework operationalizes this principle through four
components, each targeting one of the gaps above.
\textbf{(1)~\toolname{} + \datasetname{} (Gap~i).}
\toolname{} unifies 972,247 trajectories across 85K tasks from
10 open-source datasets, selects representative samples via
dynamic time warping (DTW)-based clustering, and annotates them
with process-level descriptions across ten fine-grained dimensions
(Table~\ref{tab:fine_grained_schema}). This produces
\textbf{\datasetname{}}, a human-verified corpus of 47,159
trajectories whose average instruction length increases
10.4$\times$ (from 9.3 to 96.8 words).
\textbf{(2)~\benchmarkname{} (Gap~ii).}
We curate \benchmarkname{}---500 videos with 11,631 human-reviewed
atomic facts and 1,030 VQA questions spanning all ten fine-grained
dimensions, with complementary VQA and caption tracks. All
benchmark trajectories are held out from both VLM fine-tuning and
policy training, ensuring an independent evaluation.
\textbf{(3)~\vlmname{} (Gap~ii).}
We fine-tune Qwen3.5-397B-A17B~\citep{qwen35} on \datasetname{}
to obtain \vlmname{}, a VLM specialized for robotic action
understanding that serves as a scalable annotator for new
trajectories.
\textbf{(4)~\policyname{} + training recipe (Gap~iii).}
We train \policyname{} under two action-decoding architectures
(StarVLA-OFT and StarVLA-GR00T) and systematically vary the ratio
between fine-grained (FG) and raw goal-level (Raw) instructions---
keeping trajectories, actions, and visual observations fixed while
changing only the paired language---to isolate the effect of
action-aligned supervision.

Our policy experiments yield three key findings.
First, fine-grained supervision does not harm goal-level task
success; instead, FG-only outperforms Raw-only across the
evaluated simulation settings, with the largest gain on
AlohaMix-OFT (+6.5/+4.7 points on Easy/Hard).
Second, fine-grained and raw instructions are complementary:
success follows an inverted-U trend over the FG:Raw ratio and
peaks around 1:2--1:1, reaching \textbf{86.8\%}/\textbf{82.5\%}
on AlohaMix-OFT Easy/Hard (+15.0/+11.1 over the Raw-only
baseline of 71.8\%/71.4\%).
Third, in real-world dual-arm manipulation, the FG:Raw\,$=$\,1:1
policy achieves the highest average score (\textbf{62.7}/100
vs.\ 49.9 for Raw-only), with the largest per-factor gains on
execution-sensitive attributes such as pose (+23), color (+18),
and approach direction (+18). Under identical visual scenes,
varying only the fine-grained instruction produces distinctly
different execution behaviors, directly demonstrating steerable
control.
In addition, \vlmname{} achieves the best performance among
evaluated VLMs on the held-out \benchmarkname{}, reaching
\textbf{68.2\%} VQA accuracy and \textbf{82.2\%} captioning
score under the hard setting, providing evidence that our
annotation schema captures action-relevant manipulation details.
We release the complete \framework{} suite---data pipeline,
fine-grained annotations, benchmark, model checkpoints, and
training code---to provide open foundations for steerable VLA
research.

%% file: sections/dataset.tex
\section{FineVLA Data: Construction, Benchmark, and Scalable Annotation}
\label{sec:tool}
\label{sec:roboact}

This section describes the data and annotation substrate of
\framework{}. We first unify heterogeneous robot demonstrations and
construct human-verified fine-grained action-aligned instructions.
We then build a held-out benchmark to evaluate process-level robotic
video understanding, and finally instantiate \vlmname{} as a scalable
annotator for extending the same annotation schema to new
trajectories.

\begin{figure}[!t]
  \centering
  \includegraphics[width=\linewidth]{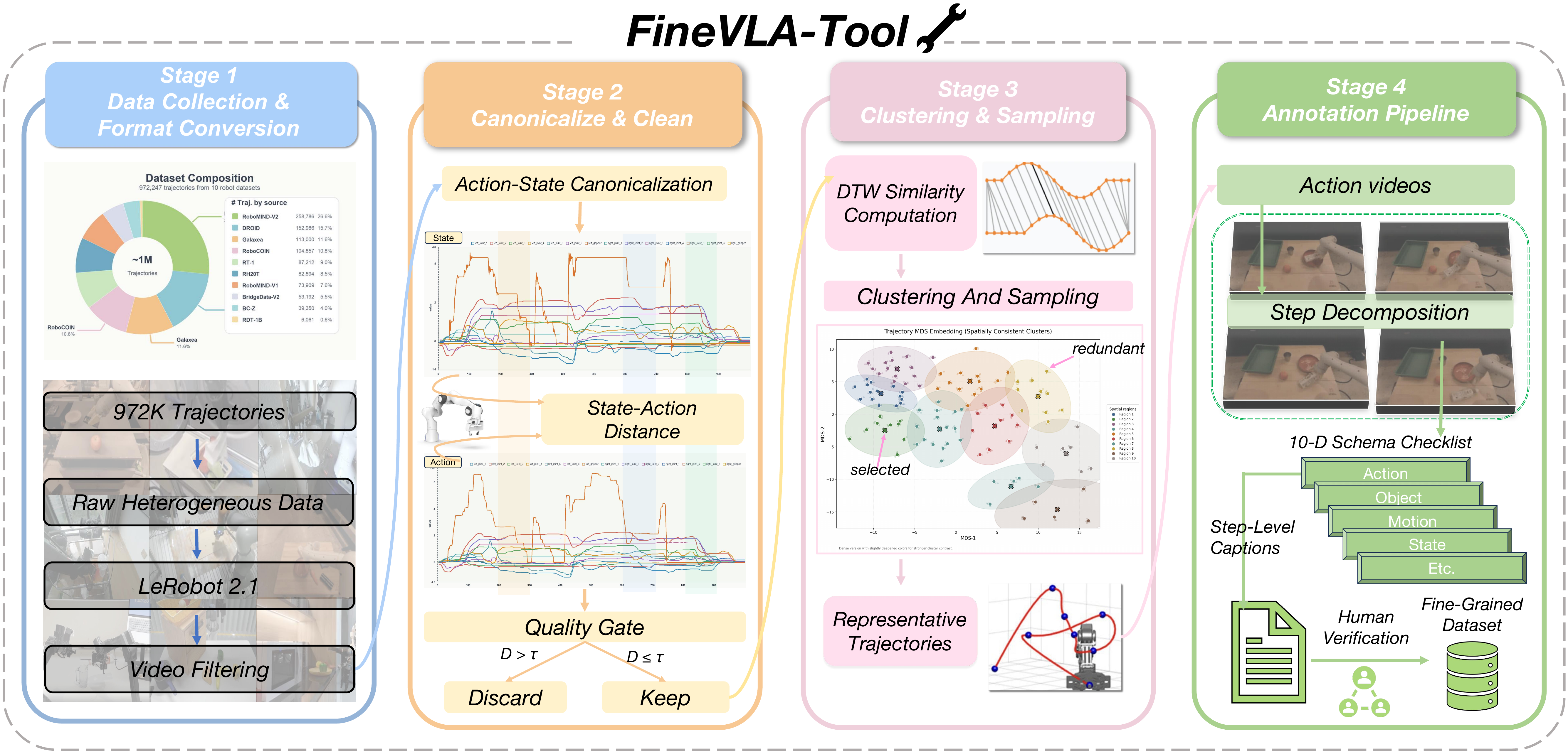}
  \vspace{-8pt}
  \caption{\textbf{Pipeline of \toolname{}.}
  \toolname{} converts large-scale heterogeneous robot
  demonstrations into action-aligned fine-grained instruction data
  through four stages.
  \textbf{Stage~1}: raw trajectories from 10 open-source robot
  datasets are converted into a unified LeRobot-style format and
  filtered to remove invalid videos.
  \textbf{Stage~2}: action and state representations are
  canonicalized across embodiments, and an action-state consistency
  quality gate removes corrupted or inconsistent trajectories.
  \textbf{Stage~3}: dynamic time warping (DTW)-based similarity
  computation and clustering identify representative trajectories,
  reducing redundancy while preserving diverse manipulation
  strategies.
  \textbf{Stage~4}: selected trajectories are decomposed into
  step-level descriptions and annotated with a ten-dimensional
  fine-grained schema, followed by human verification. The
  resulting \datasetname{} provides human-verified, process-level
  supervision for training \vlmname{} as a scalable annotator and
  \policyname{} as a steerable VLA policy.}
  \label{fig:tool_pipeline}
\end{figure}

\toolname{} converts large-scale heterogeneous robot datasets into
fine-grained, action-aligned instruction supervision
(Figure~\ref{fig:tool_pipeline}). Its design addresses three practical
bottlenecks in open robot data: (1)~inconsistent action/state formats across
datasets, (2)~heavy redundancy among demonstrations of the same task, and
(3)~sparse, task-level instruction annotations.
Starting from 972,247 trajectories across 10 source datasets, \toolname{}
produces \datasetname{}, a human-verified corpus of 47,159 representative
trajectories with fine-grained process-level supervision.

\subsection{\toolname{}: Canonicalization, Clustering, and Annotation}
\label{sec:tool_pipeline}
\label{sec:source_data}

\textbf{Data collection and format conversion
(Figure~\ref{fig:tool_pipeline}, Stage~1).}
We aggregate 972,247 trajectories from 10 open-source
datasets~\citep{bridgedata_v2,bc_z,rt1,galaxea_g0,robomind_v1,robomind_v2,robocoin,RH20T,RDT-1B,droid},
convert them to the LeRobot~2.1 format, and filter out invalid or
degenerate recordings. The full per-dataset breakdown is in
Appendix~\ref{app:tool_sources}.

\textbf{Action-state canonicalization and cleaning(Figure~\ref{fig:tool_pipeline}, Stage~2).}
Across datasets, action and state values differ in temporal reference
(absolute, relative, or delta) and kinematic representation (joint
space vs.\ end-effector space with varied rotation encodings). We
canonicalize all trajectories to absolute coordinates with normalized
quaternion rotations, then remove trajectories whose action-state DTW
distance exceeds a dataset-specific threshold, filtering out corrupted
logs or inconsistent control conventions. Details and conversion
examples are in Appendix~\ref{app:canonicalization}.

\textbf{Trajectory clustering and representative sampling(Figure~\ref{fig:tool_pipeline}, Stage~3).}
Open robot datasets contain many near-duplicate demonstrations within the same
task, often differing only in execution speed or minor spatial offsets. To
maximize annotation diversity under a fixed budget, we cluster trajectories
within each task using DTW over canonicalized action sequences, followed by
hierarchical clustering on the resulting distance matrix. We then
select high-quality representatives from each cluster according to
cluster size and trajectory quality. This reduces 972,247 raw trajectories to 47,159
representative samples while preserving diversity in manipulation strategies,
object interactions, and motion patterns. Details of the DTW formulation,
action-space normalization, frame costs, and clustering procedure are provided
in Appendix~\ref{app:tool_clustering}.

\textbf{Fine-grained multi-aspect annotation(Figure~\ref{fig:tool_pipeline}, Stage~4).}
Each selected trajectory is annotated with a ten-dimensional schema capturing
the control-relevant factors that goal-level instructions omit:
\textit{action sequence}, \textit{active actor}, \textit{target object},
\textit{initial configuration}, \textit{final configuration},
\textit{contact and approach}, \textit{trajectory and orientation},
\textit{object interaction}, \textit{failure and recovery}, and
\textit{body motion}. Detailed definitions and examples are provided in
Appendix~\ref{app:fine_grained_schema},
Table~\ref{tab:fine_grained_schema}. Annotation proceeds in two phases: we first input sampled video frames from
each trajectory into Qwen3.5-Plus~\citep{qwen35}, which decomposes the
manipulation process into temporally ordered steps and fills structured slots
for actor, target, contact region, motion path, and state change; human
annotators then review the model-generated descriptions against the original
video, correcting factual errors and verifying temporal alignment. The result is
\datasetname{}, a human-verified fine-grained instruction dataset for training
\vlmname{} and downstream controllable VLA policies.

\subsection{\datasetname{} Statistics}
\label{sec:dataset_stats}

\input{tables/dataset_stats}

Table~\ref{tab:dataset_stats} summarizes the statistics of \datasetname{}.
Fine-grained annotations dramatically increase instruction information
density compared to original coarse instructions: the average word count
per trajectory rises from 9.3 to 96.8, an approximately 10.4$\times$
increase, while covering 47 unique action verbs across all sources.
Detailed source dataset statistics are reported in
Appendix~\ref{app:tool_sources}, Table~\ref{tab:tool_source_datasets}.

\subsection{\benchmarkname{}: Fine-Grained Robotic Video Understanding Benchmark}
\label{sec:roboactbench}

\begin{figure}[t]
  \centering
  \includegraphics[width=\linewidth]{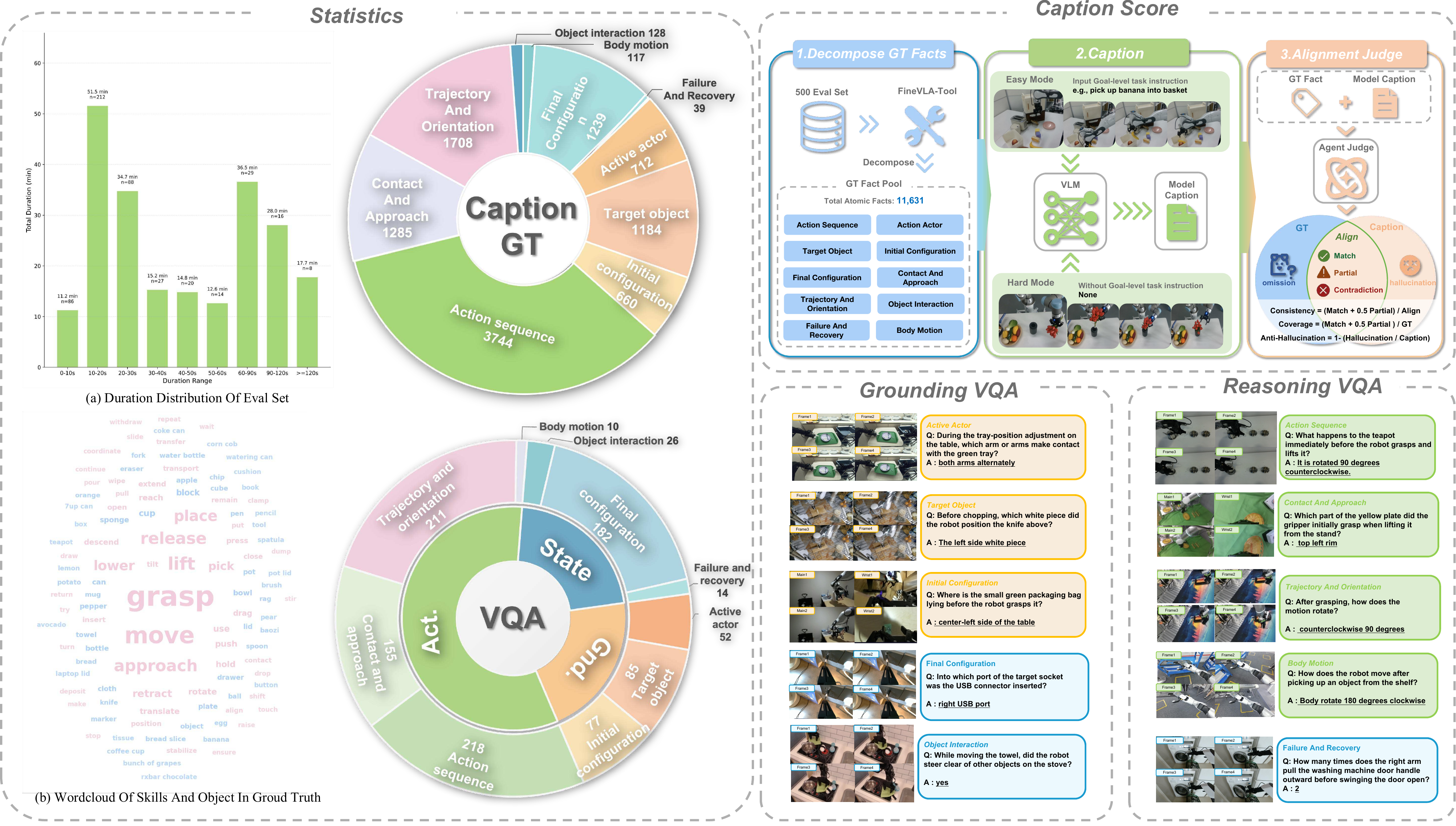}
  \caption{\textbf{Overview of \benchmarkname{}.}
  \benchmarkname{} evaluates fine-grained robotic video understanding
  through complementary VQA and captioning tracks.
  \textbf{Left}: benchmark statistics, including the video-duration
  distribution, the word cloud of manipulation skills and objects,
  and the distribution of ground-truth atomic facts across the ten
  FineVLA dimensions for captioning and VQA.
  \textbf{Top right}: the captioning track decomposes human-reviewed
  annotations into a pool of 11,631 atomic facts, asks VLMs to
  generate ordered step-level action descriptions under easy and
  hard settings, and uses an LLM judge to align model captions with
  ground-truth facts, producing Consistency, Coverage, and
  Anti-Hallucination scores.
  \textbf{Bottom right}: representative Grounding VQA and Reasoning
  VQA examples probe object/scene grounding, action/motion
  understanding, and interaction/state reasoning.
  \benchmarkname{} contains 500 held-out robot videos and 1,030 VQA
  questions across diverse embodiments, camera views, and
  manipulation scenarios.}
  \label{fig:benchmark_overview}
\end{figure}

We introduce \benchmarkname{} to evaluate whether VLMs capture
execution-level details of robot manipulation. The benchmark contains 500
videos from 10 robot datasets, covering 32 embodiments, diverse camera views,
and a wide range of manipulation tasks. Each trajectory is paired with
human-reviewed step-level annotations decomposed into 11,631 atomic facts
across ten action-relevant dimensions
(Table~\ref{tab:fine_grained_schema}), with an average of 4.3 steps and
23.3 facts per sample. All 500 benchmark trajectories are strictly disjoint
from both the \vlmname{} SFT training set and all policy-training
splits---no trajectory appears in both the 47,159 training samples
and the benchmark. Figure~\ref{fig:benchmark_overview} illustrates the benchmark statistics and structure.

\benchmarkname{} contains two complementary tracks.
The \textbf{VQA track}
(Figure~\ref{fig:benchmark_overview}, right bottom) evaluates
discriminative understanding through 1,030 questions distributed
across the same ten fine-grained dimensions used in annotation,
which are aggregated into three reporting axes:
\textbf{Entity and Scene Grounding}, \textbf{Action and Motion
Understanding}, and \textbf{Interaction and State Reasoning}
(Table~\ref{tab:vqa_dim_mapping} in
Appendix~\ref{app:vqa_dim_mapping}).
Each model receives video frames and all questions for one sample
in a single prompt, and answers are scored by
deterministic matching against ground-truth labels. The \textbf{Caption track}(Figure~\ref{fig:benchmark_overview}, right top)
evaluates generative understanding by asking models to produce action-aligned
step-level fine-grained descriptions of the manipulation process. Generated captions are then
judged by an LLM against pre-extracted ground-truth atomic facts, yielding
per-fact alignment labels (match, partial, contradiction, omission, hallucination)
that are aggregated into Consistency, Coverage, and Anti-Hallucination metrics.
Two settings are evaluated: \emph{easy}, where the original task instruction is
provided, and \emph{hard}, where the model must infer the process from visual
observations alone. Full prompt templates and the evaluation protocol are provided
in Appendix~\ref{sec:app_bench_prompts}.

\subsection{\vlmname{}: Scalable Fine-Grained Annotator}
\label{sec:roboactvlm}

While \datasetname{} is human verified, scaling the same annotation
schema to future robot trajectories still requires a
robotics-specialized annotator. General-purpose VLMs often miss
execution-level details such as contact regions, approach directions,
grasp types, motion paths, and object state transitions, leading to
substantial human correction cost.

We therefore fine-tune Qwen3.5-397B-A17B~\citep{qwen35} on the
human-verified \datasetname{} to obtain \vlmname{}. Given a robot
manipulation video, \vlmname{} generates temporally action-aligned 
step-level fine-grained descriptions covering the ten fine-grained
dimensions (Table~\ref{tab:fine_grained_schema}). The model serves as a
scalable annotator for future data expansion; all policy experiments
in this paper use \toolname{}-generated and human-verified
annotations, rather than \vlmname{}-generated labels.
Importantly, \vlmname{} is not used to generate the supervision
for our policy experiments; it is trained to support future
scalable annotation and open-source reproducibility.
Full training details are provided in
Appendix~\ref{sec:app_vlm_training}; its annotation quality is
evaluated on the held-out \benchmarkname{} in
Section~\ref{sec:benchmark_results}.

%% file: tables/dataset_stats.tex
\begin{table}[t]
  \caption{\textbf{\datasetname{} statistics.} Fine-grained
  annotations dramatically increase instruction information density
  compared to original coarse instructions across all data sources.}
  \label{tab:dataset_stats}
  \centering
  \small
  \begin{tabular}{lrrrrr}
    \toprule
    Source & Trajectories & Steps
      & Avg.\ Words (Coarse) & Avg.\ Words (FG)
      & Density $\uparrow$ \\
    \midrule
    BridgeData-V2  & 4,958  & 21,554  &  10.1 &  61.7 &  6.1$\times$ \\
    BC-Z           & 1,513  &  5,313  &   5.2 &  51.2 &  9.8$\times$ \\
    RT-1           & 5,232  & 22,023  &   6.8 &  61.4 &  9.1$\times$ \\
    Galaxea        & 2,834  & 18,484  &   4.7 & 219.9 & 47.1$\times$ \\
    RoboMIND-V1    & 4,605  & 20,341  &   8.6 &  72.8 &  8.5$\times$ \\
    RoboMIND-V2    & 7,119  & 39,166  &   6.6 &  98.8 & 14.9$\times$ \\
    RoboCOIN       & 8,513  & 43,926  &  16.1 & 122.6 &  7.6$\times$ \\
    RH20T          & 1,387  &  5,560  &   7.9 &  92.1 & 11.7$\times$ \\
    RDT            & 1,275  &  8,437  &  16.9 & 114.0 &  6.7$\times$ \\
    DROID          & 9,723  & 35,802  &   8.0 &  90.9 & 11.3$\times$ \\
    \midrule
    \textbf{Total} & \textbf{47,159} & \textbf{220,606}
      & 9.3 & 96.8 & 10.4$\times$ \\
    \bottomrule
  \end{tabular}
\end{table}

%% file: sections/method.tex
\section{Training Fine-Grained VLA Policies}
\label{sec:FineVLA}

With human-verified fine-grained instructions fixed, we now study how
they should be used to train steerable VLA policies. Our goal is not
to propose a new policy architecture, but to isolate the effect of
action-aligned language supervision. We therefore keep
actions, and visual observations fixed, and vary only the instruction
paired with each trajectory.

\subsection{\policyname{} Architecture}
\label{sec:policy_arch}

We instantiate \policyname{} under multiple action-decoding frameworks to
isolate the effect of instruction supervision from architectural choices.
Rather than proposing a new architecture, we adopt two existing frameworks
implemented in the StarVLA
codebase~\citep{community2026starvlalegolikecodebasevisionlanguageaction},
both built on a shared \qwenvl{} vision-language backbone.

\textbf{StarVLA-OFT} attaches a lightweight MLP regression head that reads
the hidden states of predefined action tokens and predicts continuous action
chunks in parallel with an L1 objective, following OpenVLA-OFT.
\textbf{StarVLA-GR00T} adopts a dual-system design where the VL backbone
serves as System~2 for slow reasoning and a DiT-based flow-matching module
serves as System~1 for continuous action generation, consistent with
GR00T N1.5. Both variants produce multi-step action chunks and share the same
visual observations and language inputs; only the action decoding differs.
Using two architectures lets us verify that the benefits of fine-grained
supervision are architecture-independent.

\subsection{Training Data Mixtures}
\label{sec:instruction_mixing}

To isolate the effect of language supervision, we construct two
parallel training datasets from the same source trajectories.
The \textbf{FG dataset} contains the representative trajectories
selected by \toolname{}, each paired with its fine-grained
process-level instruction (1,275 trajectories for RDT; 5,872 for
AlohaMix). The \textbf{Raw dataset} contains \emph{all} source
trajectories, each paired with its original goal-level
instruction (6,061 for RDT; 86,662 for AlohaMix). AlohaMix is an
ALOHA-compatible dual-arm mixture aggregated from RDT, RoboCOIN,
RoboMIND-V1.0, and RoboMIND-V2.0, containing 86,662 episodes
across 598 tasks (Table~\ref{tab:alohamix_composition} in
Appendix~\ref{sec:app_policy_datasets}). We restrict the mixture
to a single embodiment class to avoid cross-embodiment confounds.
Trajectories that appear in both datasets share identical action
labels and visual observations; only the paired language
instruction differs.

The FG:Raw ratio controls the probability of drawing from each
dataset at every training step, and therefore determines the
relative number of training iterations that use fine-grained
versus raw instructions. For example, FG:Raw~$=$~2:1 means the FG
dataset is sampled with twice the weight of the Raw dataset, so
approximately two-thirds of training steps use a fine-grained
instruction and one-third use a raw instruction. Under Raw-only,
training draws exclusively from the Raw dataset; under FG-only,
exclusively from the FG dataset.

We compare seven configurations: Raw-only, FG:Raw~$=$~1:4, 1:2,
1:1, 2:1, 4:1, and FG-only. We study three (dataset, framework)
combinations---RDT-OFT, RDT-GR00T, and AlohaMix-OFT---to
control for architecture and data-scale effects. This design
isolates the effect of action-aligned language supervision from
changes in data scale, embodiment, or action distribution.

%% file: sections/experiment.tex
\section{Experiments}
\label{sec:policy}

We evaluate \framework{} along three axes:
(1)~whether \vlmname{} captures fine-grained robotic action details
(\benchmarkname{}, Section~\ref{sec:benchmark_results}),
(2)~whether fine-grained supervision improves policy learning in
simulation (RoboTwin, Section~\ref{sec:policy_robotwin}), and
(3)~whether the resulting policies exhibit steerable control on
real-world dual-arm manipulation
(Section~\ref{sec:policy_real}).

\subsection{Experimental Setup}
\label{sec:exp_setup}

\paragraph{Evaluation benchmark.}
We evaluate \framework{} on three complementary evaluation
protocols that measure robotic video understanding, simulated
policy learning, and physical steerable control.

\textbf{(1)~\benchmarkname{}.}
\benchmarkname{} (Section~\ref{sec:roboactbench}) evaluates
whether \vlmname{} captures fine-grained robotic action details.
We compare \vlmname{} with five strong general-purpose VLMs on
both VQA and captioning tracks. The VQA track reports overall
and dimension-wise accuracy across the ten FineVLA dimensions,
while the captioning track scores ordered step-level action
descriptions using Consistency, Coverage, and Anti-Hallucination
metrics.

\textbf{(2)~RoboTwin Simulation Evaluation.}
RoboTwin~\citep{robotwin} evaluates simulated bimanual
manipulation. We test the seven FG:Raw instruction ratios defined
in Section~\ref{sec:instruction_mixing} across three controlled
policy settings: RDT-OFT, RDT-GR00T, and AlohaMix-OFT. Policies
are evaluated on the official Easy and Hard splits, with
20 episodes per task.

\textbf{(3)~Real-world Steerability Evaluation.}
We design this self-contained real-world benchmark on a Cobot
Magic dual-arm platform to measure language-conditioned
controllability. Unlike broad robustness benchmarks that vary
scenes, objects, or lighting, our suite isolates instruction
following: for each instruction-sensitive task family, paired
variants use the same object set and nearly identical initial
scene layout while changing only one language-specified control
factor. The suite includes two general manipulation tasks, five
in-distribution instruction-sensitive task families (each
comprising a paired variant) covering object color, object pose,
approach direction, rotation direction, and active arm, and one
out-of-distribution active-arm--target binding probe. Each paired
variant is evaluated over 10 trials and scored with a
partial-completion rubric normalized to a 0--100 scale.
Additional hardware and inference details are reported in
Appendix~\ref{sec:app_real_setup}.

\paragraph{Training setup.}
We train three policy configurations---\textbf{RDT-OFT},
\textbf{RDT-GR00T}, and \textbf{AlohaMix-OFT}---to decouple
architecture and data-mixture effects. RDT-OFT and RDT-GR00T use
the same RDT pretraining data with different action decoders,
while RDT-OFT and AlohaMix-OFT use the same OFT decoder with
different pretraining mixtures. We pretrain each configuration
for 100k steps on 64 A100 GPUs with per-device batch size 8 and
global batch size 512.

For RoboTwin evaluation, we fine-tune the pretrained checkpoints
on the union of the Clean and Random training sets, containing
27,500 trajectories and 6,075,103 transitions, for 100k steps on
8 A100 GPUs with global batch size 128. The FG:Raw instruction
mixture is applied during this fine-tuning stage; pretraining
uses the original instruction format of each source dataset.

For real-world evaluation, we further fine-tune the corresponding
simulation checkpoint for each instruction-mixture setting on 50
demonstrations per task from 12 tabletop tasks, for 600
demonstrations in total, collected on the Cobot Magic dual-arm
platform. Full optimizer, hardware, batch-size, and
training-step configurations are reported in
Appendix~\ref{app:policy_setup},
Table~\ref{tab:policy_training_configs}.

\subsection{RoboFine-Bench Results}
\label{sec:benchmark_results}

\input{tables/benchmark}

Tables~\ref{tab:benchmark_vqa} and~\ref{tab:benchmark_caption} compare
\vlmname{} with five strong general-purpose VLMs on \benchmarkname{}.
A more detailed analysis of the benchmark results is provided in
Appendix~\ref{app:benchmark_detailed_analysis}.

\textbf{VQA results.}
\vlmname{} achieves \textbf{68.2\%} overall accuracy, outperforming the
strongest general-purpose baseline, GPT-5.4, by \textbf{8.0} absolute
points. The largest gain appears on Action and Motion Understanding
(\textbf{75.7\%} vs.\ \textbf{64.6\%}), indicating that fine-grained
supervision improves execution-level reasoning beyond scene recognition.
Compared with its base model Qwen3.5-Plus (i.e., Qwen3.5-397B-A17B), SFT on \datasetname{} improves
overall accuracy from \textbf{55.9\%} to \textbf{68.2\%}, with consistent gains
across grounding, action, and state reasoning.

\textbf{Caption results.}
\vlmname{} also leads the caption track. In the easy setting, where the task
instruction is provided, it obtains the best Overall, Consistency, and Coverage
scores. In the hard setting, where the model must infer the manipulation process
from video alone, \vlmname{} ranks first on all four metrics and improves
Overall from the strongest baseline score of \textbf{78.0\%} to
\textbf{82.2\%}. This setting is especially important because it measures
whether the model captures the execution process rather than relying on
task-level language priors. Token and latency statistics are reported in
Appendix~\ref{app:caption_cost}.

\begin{figure}[t]
  \centering
  \begin{subfigure}[t]{0.49\linewidth}
    \centering
    \includegraphics[width=\linewidth]{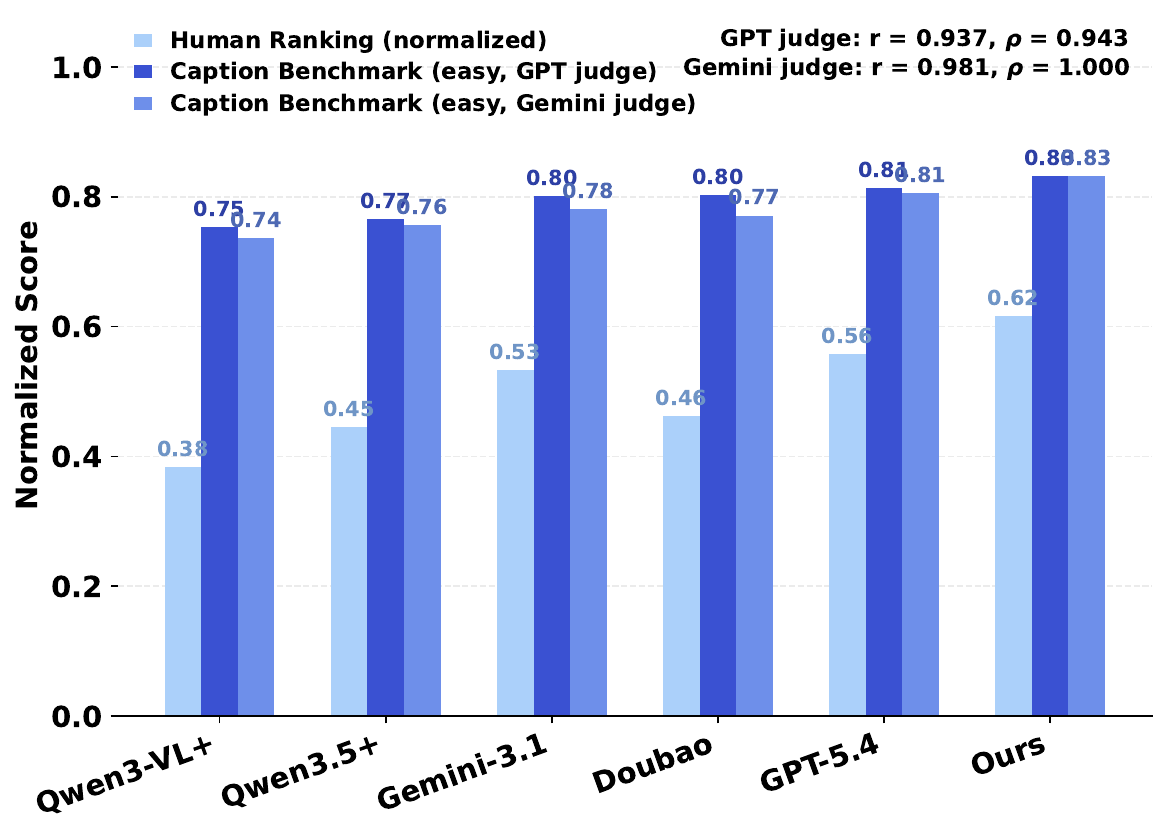}
    \caption{Comparison on easy mode.}
  \end{subfigure}
  \hfill
  \begin{subfigure}[t]{0.49\linewidth}
    \centering
    \includegraphics[width=\linewidth]{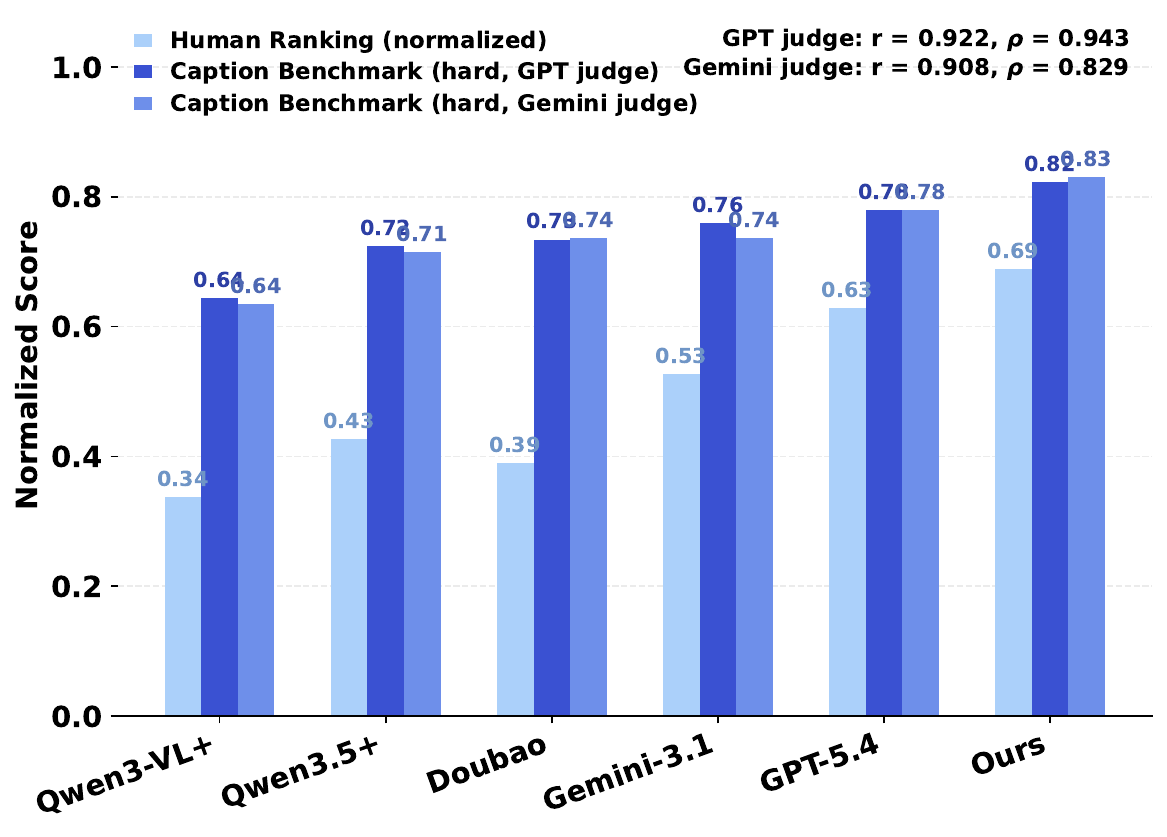}
    \caption{Comparison on hard mode.}
  \end{subfigure}
  \caption{\textbf{Correlation between benchmark caption scores and human ranking.}
  We recruit 10 human raters to rank the six models on the 500 benchmark videos,
  and average the resulting subjective scores. Human ranks are normalized from
  the 1--6 range to $[0,1]$, while benchmark caption Overall scores are normalized
  from 0--100 to $[0,1]$.}
  \label{fig:caption_human_alignment}
\end{figure}

\textbf{Benchmark validity.}
The caption ranking is robust to the choice of alignment judge: replacing
GPT-5.4-Pro with Gemini-3.1-Pro preserves the same strongest model in both
easy and hard settings, with only small changes in absolute scores
(Appendix~\ref{app:caption_judge_robustness}). The automatic scores also align
strongly with human preference. As shown in
Figure~\ref{fig:caption_human_alignment}, the correlation between benchmark
Overall scores and the rankings from 10 human raters is high in both settings
(easy: Pearson \textbf{0.937}, Spearman $\rho$ \textbf{0.943}; hard: Pearson
\textbf{0.922}, Spearman $\rho$ \textbf{0.943}).

These results provide evidence that \vlmname{} can produce dense,
action-aligned robotic descriptions, and that the proposed
annotation schema captures execution-level manipulation details.
Importantly, the fine-grained supervision used for policy training
is produced by \toolname{} with human verification, rather than by
\vlmname{}. Thus, \vlmname{} is evaluated here as a scalable
annotator for future data expansion.

\subsection{RoboTwin Simulation Results}
\label{sec:policy_robotwin}

\input{tables/robotwin}

We evaluate on RoboTwin~\citep{robotwin}, a simulation benchmark for
bimanual manipulation, and report success rate on its official
\textbf{Easy} and \textbf{Hard} splits. Each policy is evaluated over
20 episodes per task and averaged to produce the per-split score.
Table~\ref{tab:robotwin} reports results across three (dataset,
framework) combinations: RDT-OFT, RDT-GR00T, and AlohaMix-OFT.
Note that AlohaMix is approximately 13$\times$ larger than RDT in
episode count, enabling a controlled study of data-scale effects.

Table~\ref{tab:robotwin} shows two main results.
First, fine-grained supervision does not harm goal-level task
success: FG-only improves over Raw-only across all evaluated
settings, with gains of +1.4/+2.0 on RDT-OFT (Easy/Hard),
+7.0/+8.1 on RDT-GR00T, and +6.5/+4.7 on AlohaMix-OFT.
Second, fine-grained and raw instructions are complementary:
as the FG proportion increases from 0\% to 100\%, success rate
follows a consistent inverted-U trend across all three settings,
peaking around FG\,:\,Raw\,$=$\,1\,:\,2 to 1\,:\,1. The best
setting, FG\,:\,Raw\,$=$\,1\,:\,1, reaches
\textbf{86.8\%}/\textbf{82.5\%} on AlohaMix-OFT Easy/Hard,
a gain of +15.0/+11.1 over the Raw-only baseline
(71.8\%/71.4\%). Both conclusions hold across all three
(dataset, framework) combinations, regardless of action-decoding
architecture (OFT vs.\ GR00T) and pretraining data scale (RDT
vs.\ AlohaMix). We analyze this trend and its mechanism in
Section~\ref{sec:complementary}.

\subsection{Real-World Steerability Results}
\label{sec:policy_real}

\begin{figure}[!t]
  \centering
  \includegraphics[width=\linewidth]{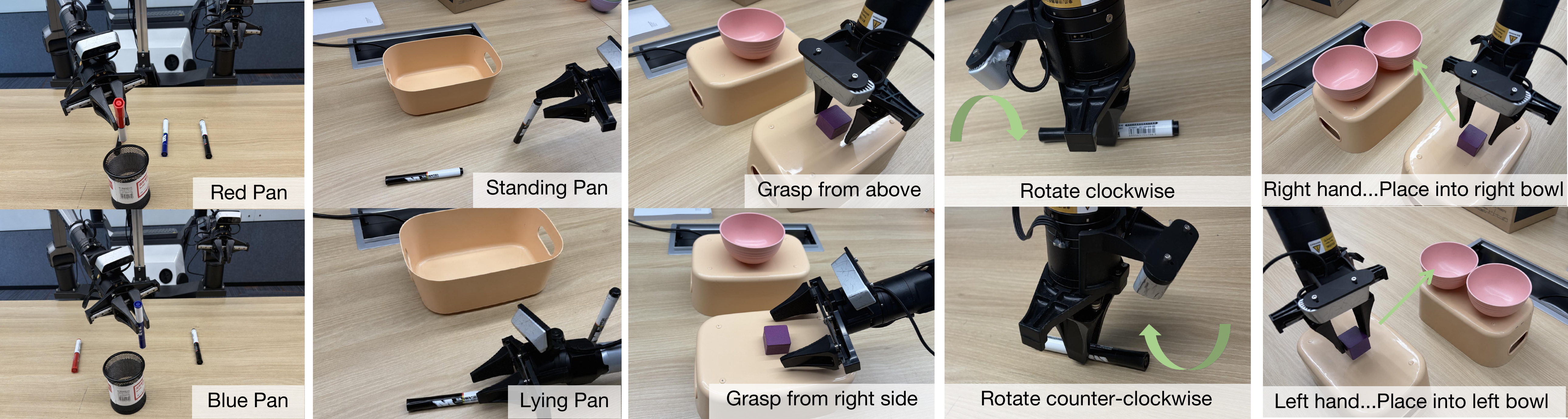}
  \vspace{-8pt}
  \caption{\textbf{Paired real-world evaluation.} Each column shows
  one control factor under the same visual scene with two language
  variants. From left to right: Color (red/blue), Pose
  (standing/lying), Approach (above/side), Rotation
  (clockwise/counterclockwise), Arm (right/left).}
  \label{fig:real_world_tasks}
\end{figure}

\input{tables/real_robot}

We evaluate physical steerable control on our self-designed
Real-world Steerability Suite
(Figure~\ref{fig:real_world_tasks}). This benchmark is designed
to isolate language-conditioned controllability: paired task
variants share nearly the same visual scene but require different
behaviors according to the instruction, such as choosing a
different object color, object pose, approach direction,
rotation direction, or active arm.

Table~\ref{tab:real} shows two main results.
First, fine-grained supervision improves steerable control---a
conclusion that simulation benchmarks alone cannot provide.
FG\,:\,Raw\,$=$\,1\,:\,1 improves every instruction-sensitive
factor over Raw-only: Color (22\,$\to$\,\textbf{40}),
Pose (24\,$\to$\,\textbf{47}),
Approach (60\,$\to$\,\textbf{78}),
Rotate (76\,$\to$\,\textbf{86}), and
Arm (60\,$\to$\,\textbf{64}). The largest gains appear on
factors invisible to goal-level language---Pose (+23), Color
and Approach (+18 each)---precisely the execution choices that
raw instructions leave unspecified.
Second, consistent with the simulation findings, fine-grained
and raw instructions are complementary in the real world. The
in-domain score follows a clear inverted-U trend across all
seven settings (49.9 $\to$ 53.9 $\to$ 61.1 $\to$
\textbf{62.7} $\to$ 58.7 $\to$ 57.4 $\to$ 54.4), peaking at
FG\,:\,Raw\,$=$\,1\,:\,1 (\textbf{62.7}/100), which outperforms
both Raw-only (49.9) and FG-only (54.4). On the two general
manipulation tasks, Clean Table (72 $\to$ \textbf{84}) and Stack
Block (35 $\to$ \textbf{40}), mixed supervision also matches or
exceeds Raw-only, indicating that process-level language does not
interfere with routine execution. The OOD actor-target binding
remains challenging, but the mixed model improves from 0 to
10/100, suggesting partial factor-level generalization. We
analyze factor-level controllability and failure modes in
Section~\ref{sec:language_critical}.

%% file: tables/benchmark.tex
{\setlength{\floatsep}{6pt}
\setlength{\textfloatsep}{8pt}
\setlength{\abovecaptionskip}{4pt}
\setlength{\belowcaptionskip}{2pt}
\begin{table}[!t]
  \caption{\textbf{VQA benchmark results on \benchmarkname{} (\%).}
  We report overall VQA accuracy together with all ten fine-grained capability
  dimensions. \textbf{AA}: Active Actor; \textbf{TO}: Target Object;
  \textbf{IC}: Initial Configuration; \textbf{AS}: Action Sequence;
  \textbf{C\&A}: Contact \& Approach; \textbf{T\&O}: Trajectory \&
  Orientation; \textbf{BM}: Body Motion; \textbf{OI}: Object Interaction;
  \textbf{FC}: Final Configuration; \textbf{F\&R}: Failure \& Recovery.
  Best value per column is \textbf{bold}.}
  \label{tab:benchmark_vqa}
  \centering
  \scriptsize
  \setlength{\tabcolsep}{2pt}
  \renewcommand{\arraystretch}{1.05}
  \begin{tabular*}{\linewidth}{@{\extracolsep{\fill}}l!{\hspace{2pt}\vrule\hspace{2pt}}c!{\hspace{2pt}\vrule\hspace{2pt}}ccc!{\hspace{2pt}\vrule\hspace{2pt}}cccc!{\hspace{2pt}\vrule\hspace{2pt}}ccc@{}}
    \toprule
    \multirow{2}{*}{Model}
      & \multirow{2}{*}{Overall$\uparrow$}
      & \multicolumn{3}{c}{Gnd.$\uparrow$}
      & \multicolumn{4}{c}{Act.$\uparrow$}
      & \multicolumn{3}{c}{State$\uparrow$} \\
    \cmidrule(lr){3-5} \cmidrule(lr){6-9} \cmidrule(lr){10-12}
      &  & AA & TO & IC & AS & C\&A & T\&O & BM & OI & FC & F\&R \\
    \midrule
    Qwen3-VL-Plus       & 47.7 & 57.7 & 47.1 & 44.2 & 56.0 & 45.2 & 46.9 & 60.0 & 46.2 & 39.6 & 42.9 \\
    Qwen3.5-Plus        & 55.9 & 73.1 & 60.0 & 58.4 & 56.6 & 49.4 & 53.8 & 80.0 & 38.5 & 57.1 & 42.9 \\
    Doubao-Seed-2.0-Pro & 58.5 & 63.5 & 55.3 & 53.2 & 62.4 & 49.7 & 58.8 & 70.0 & 53.8 & 64.3 & 50.0 \\
    Gemini-3.1-Pro      & 59.6 & \textbf{84.6} & 60.0 & 53.2 & 65.1 & 58.7 & 51.7 & 80.0 & 50.0 & 58.8 & 57.1 \\
    GPT-5.4             & 60.2 & \textbf{84.6} & 60.0 & 49.4 & 64.7 & 60.7 & 53.1 & 80.0 & \textbf{61.5} & 59.9 & 50.0 \\
    \midrule
    \vlmname{} (Ours)   & \textbf{68.2} & 82.7 & \textbf{65.9} & \textbf{68.8} & \textbf{70.6} & \textbf{69.0} & \textbf{63.0} & \textbf{100.0} & \textbf{61.5} & \textbf{65.4} & \textbf{78.6} \\
    \bottomrule
  \end{tabular*}
\end{table}

\begin{table}[!t]
  \caption{\textbf{Caption benchmark results on \benchmarkname{} (\%).}
  We report caption quality under two settings: \textbf{easy}, where the
  original task instruction is provided, and \textbf{hard}, where the model
  must infer the manipulation process from vision alone. \textbf{Cons.}:
  Consistency; \textbf{Cov.}: Coverage; \textbf{A-Hal.}:
  Anti-Hallucination. Best value per column is \textbf{bold}.}
  \label{tab:benchmark_caption}
  \centering
  \footnotesize
  \renewcommand{\arraystretch}{1.05}
  \begin{tabular*}{\linewidth}{@{\extracolsep{\fill}}lcccccccc@{}}
    \toprule
    \multirow{2}{*}{Model}
      & \multicolumn{4}{c}{Easy}
      & \multicolumn{4}{c}{Hard} \\
    \cmidrule(lr){2-5} \cmidrule(lr){6-9}
      & Overall$\uparrow$ & Cons.$\uparrow$ & Cov.$\uparrow$ & A-Hal.$\uparrow$ & Overall$\uparrow$ & Cons.$\uparrow$ & Cov.$\uparrow$ & A-Hal.$\uparrow$ \\
    \midrule
    Qwen3-VL-Plus       & 75.4 & 75.2 & 58.2 & 92.8 & 64.4 & 67.4 & 54.3 & 71.6 \\
    Qwen3.5-Plus        & 76.6 & 75.3 & 59.1 & 95.5 & 72.4 & 71.0 & 55.1 & 91.2 \\
    Doubao-Seed-2.0-Pro & 80.2 & 78.5 & 68.2 & 93.8 & 73.4 & 72.4 & 63.7 & 84.0 \\
    Gemini-3.1-Pro      & 80.1 & 79.9 & 62.7 & \textbf{97.7} & 75.9 & 75.7 & 58.5 & 93.4 \\
    GPT-5.4             & 81.4 & 79.5 & 72.1 & 92.5 & 78.0 & 73.8 & 66.8 & 93.4 \\
    \midrule
    \vlmname{} (Ours)   & \textbf{83.2} & \textbf{82.1} & \textbf{72.7} & 94.8 & \textbf{82.2} & \textbf{80.4} & \textbf{71.5} & \textbf{94.8} \\
    \bottomrule
  \end{tabular*}
\end{table}
}

%% file: tables/robotwin.tex
{\setlength{\textfloatsep}{8pt}
\setlength{\abovecaptionskip}{4pt}
\setlength{\belowcaptionskip}{2pt}
\begin{table}[!t]
  \caption{\textbf{RoboTwin simulation success rate (\%).}
    We compare three training settings (RDT-OFT, RDT-GR00T, and
    AlohaMix-OFT) under seven FG:Raw instruction ratios. Easy/Hard
    follow the official RoboTwin splits. Best value per column is
    \textbf{bold}.}
  \label{tab:robotwin}
  \centering
  \small
  \renewcommand{\arraystretch}{1.05}

  \begin{tabular*}{\linewidth}{@{\extracolsep{\fill}}lcccccc@{}}
    \toprule
    \multirow{2}{*}{\textbf{FG:Raw}}
      & \multicolumn{2}{c}{\textbf{RDT-OFT}}
      & \multicolumn{2}{c}{\textbf{RDT-GR00T}}
      & \multicolumn{2}{c}{\textbf{AlohaMix-OFT}} \\
    \cmidrule(lr){2-3} \cmidrule(lr){4-5}
    \cmidrule(lr){6-7}
      & Easy$\uparrow$ & Hard$\uparrow$ & Easy$\uparrow$ & Hard$\uparrow$ & Easy$\uparrow$ & Hard$\uparrow$ \\
    \midrule
    Raw-only                    & 61.5 & 60.0 & 55.1 & 53.4 & 71.8 & 71.4 \\
    FG\,:\,Raw $=$ 1\,:\,4      & 68.2 & 66.5 & 58.2 & 55.7 & 75.3 & 74.3 \\
    FG\,:\,Raw $=$ 1\,:\,2      & \textbf{74.1} & 72.1 & 61.7 & 60.9 & 82.8 & 78.6 \\
    FG\,:\,Raw $=$ 1\,:\,1      & 73.9 & \textbf{72.4} & \textbf{69.4} & \textbf{68.2} & \textbf{86.8} & \textbf{82.5} \\
    FG\,:\,Raw $=$ 2\,:\,1      & 70.4 & 68.3 & 65.9 & 63.1 & 80.9 & 79.3 \\
    FG\,:\,Raw $=$ 4\,:\,1      & 68.6 & 67.5 & 64.9 & 63.2 & 79.5 & 78.5 \\
    FG-only                     & 62.9 & 62.0 & 62.1 & 61.5 & 78.3 & 76.1 \\
    \bottomrule
  \end{tabular*}
\end{table}
}

%% file: tables/real_robot.tex
\begin{table*}[t]
  \caption{\textbf{Real-world scores on a 100-point scale.}
    All models use StarVLA-OFT pretrained on AlohaMix (100k fine-tuning steps).
    Each trial is scored by manually checking ordered subgoals;
    a completed subgoal receives proportional credit, and the final score is normalized to 100.
    \emph{Avg\,(ID)} averages the seven non-OOD tasks
    (two general $+$ five instruction-sensitive);
    \emph{Avg\,(All)} additionally includes the OOD probe ($\dagger$).
    \textbf{Bold} indicates the best score per column.
    Task descriptions and control factors are listed in Table~\ref{tab:real_tasks}.}
  \label{tab:real}
  \centering
  \small
  \setlength{\tabcolsep}{4.5pt}
  \renewcommand{\arraystretch}{1.12}
  \begin{tabular}{@{}l ccccccc @{\hspace{6pt}} c @{\hspace{2pt}} cc @{}}
    \toprule
    \multirow{2}{*}{\textbf{Supervision}} &
    \multicolumn{7}{c}{\textbf{In-Distribution Tasks}} &
    \textbf{OOD} &
    \multicolumn{2}{c}{\textbf{Average}} \\
    \cmidrule(lr){2-8} \cmidrule(lr){9-9} \cmidrule(lr){10-11}
    &
    \shortstack[c]{\textbf{Clean}\\\textbf{Table}} &
    \shortstack[c]{\textbf{Stack}\\\textbf{Block}} &
    \textbf{Color} &
    \textbf{Pose} &
    \textbf{Approach} &
    \textbf{Rotate} &
    \textbf{Arm} &
    \textbf{L$\to$R}$^\dagger$ &
    \textbf{(ID)} & \textbf{(All)} \\
    \midrule
    Raw-only
      & 72 & 35 & 22 & 24 & 60 & 76 & 60 & 0
      & 49.9 & 43.6 \\
    FG\,:\,Raw $=$ 1\,:\,4
      & 76 & 36 & 28 & 32 & 65 & 79 & 61 & 0
      & 53.9 & 47.1 \\
    FG\,:\,Raw $=$ 1\,:\,2
      & 79 & 39 & 36 & \textbf{48} & 76 & \textbf{87} & 63 & 5
      & 61.1 & 54.1 \\
    FG\,:\,Raw $=$ 1\,:\,1
      & \textbf{84} & \textbf{40} & \textbf{40} & 47 & \textbf{78} & 86 & \textbf{64} & \textbf{10}
      & \textbf{62.7} & \textbf{56.1} \\
    \midrule
    FG\,:\,Raw $=$ 2\,:\,1
      & 80 & 38 & 34 & 42 & 72 & 83 & 62 & 5
      & 58.7 & 52.0 \\
    FG\,:\,Raw $=$ 4\,:\,1
      & 74 & 37 & 31 & 43 & 72 & 83 & 62 & 5
      & 57.4 & 50.9 \\
    FG-only
      & 70 & 35 & 25 & 41 & 70 & 80 & 60 & 0
      & 54.4 & 47.6 \\
    \bottomrule
  \end{tabular}
  \vspace{3pt}
  \begin{minipage}{\linewidth}
    \footnotesize
    $^\dagger$~\emph{L\,$\to$\,R}: use left hand to place into
    right bowl; unseen actor-target combination (OOD compositional
    probe). Control factors are detailed in
    Appendix~\ref{sec:app_real_setup}.
  \end{minipage}
\end{table*}

%% file: sections/analysis.tex
\section{Analysis}
\label{sec:analysis}

This section analyzes \emph{why} fine-grained supervision improves
performance, how it should be mixed with raw goal-level instructions,
and which control factors benefit most from action-aligned language.

\subsection{Fine-Grained Supervision Does Not Sacrifice Goal-Level Success}
\label{sec:fg_universal}

A natural concern is that fine-grained instructions may
over-specify execution details and distract the policy from
completing the high-level goal. Our results suggest the opposite.
In RoboTwin (Table~\ref{tab:robotwin}), FG-only improves over
Raw-only across all three (dataset, framework) combinations,
with gains ranging from +1.4/+2.0 (RDT-OFT, Easy/Hard) to
+7.0/+8.1 (RDT-GR00T) and +6.5/+4.7 (AlohaMix-OFT). In the
real-world evaluation (Table~\ref{tab:real}), mixed supervision also matches or exceeds Raw-only on the two general
manipulation tasks (Clean Table, Stack Block), where no
fine-grained control factor is explicitly tested. The pattern
holds regardless of decoder architecture (OFT vs.\ GR00T),
pretraining data scale (RDT vs.\ AlohaMix), and environment
(simulation vs.\ physical), indicating that process-level
language provides additional action constraints without
sacrificing goal-level task completion.

\subsection{Raw and Fine-Grained Supervision Are Complementary}
\label{sec:complementary}

\begin{figure}[t]
  \centering
  \includegraphics[width=0.75\linewidth]{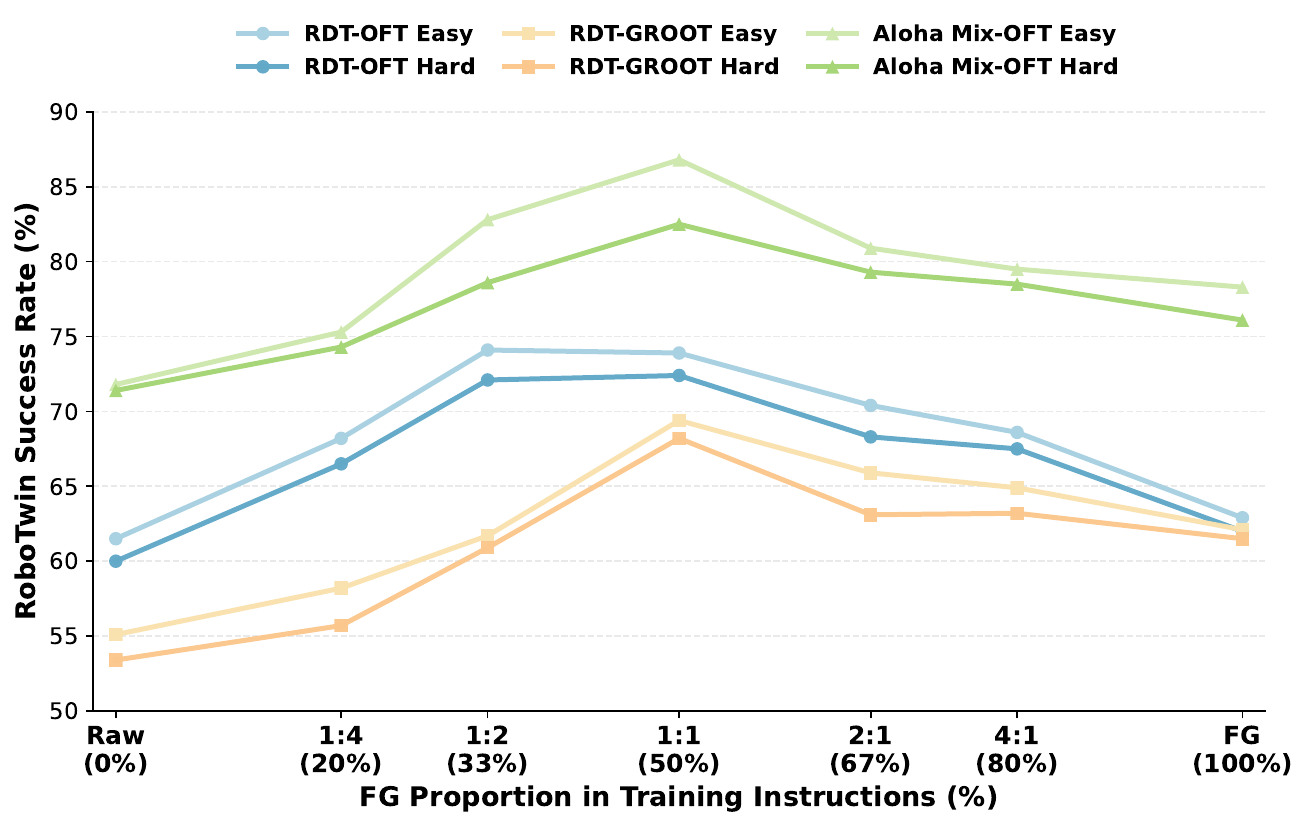}
  \caption{\textbf{RoboTwin mixing-ratio curves.} Performance
  peaks around FG\,:\,Raw $=$ 1\,:\,2 to 1\,:\,1 across all
  settings, yielding a consistent inverted-U trend.}
  \label{fig:robotwin_mixing_curve}
\end{figure}

Although FG-only outperforms Raw-only, the \emph{best} performance is achieved
by mixing both supervision types. As the FG proportion increases from 0\% to
100\%, success rate traces a clear inverted-U curve in all three RoboTwin
settings, peaking around FG\,:\,Raw\,$=$\,1\,:\,2 to 1\,:\,1
(Figure~\ref{fig:robotwin_mixing_curve}). The same trend transfers to the real
world (Table~\ref{tab:real}):
FG\,:\,Raw\,$=$\,1\,:\,1 achieves the highest in-domain score
(62.7/100), outperforming both Raw-only (49.9) and FG-only (54.4).

We attribute this inverted-U to the complementary roles of the
two instruction types. Raw instructions preserve compact goal
semantics---\emph{what} task should be completed---while
fine-grained instructions expose execution
constraints---\emph{how} the task should be performed.
Under \emph{Raw-only}, execution-level choices (arm, approach,
rotation) are left to implicit co-occurrence statistics.
Under \emph{FG-only}, the policy has explicit process-level
guidance, but losing goal-level abstractions may weaken
generalization to instruction phrasings not seen during training.
In addition, FG descriptions are longer and more
distributionally different from natural user commands, so
training exclusively on FG language may reduce exposure to
compact goal-level task phrasing.
Under \emph{Mixed} supervision, the policy simultaneously learns
task semantics from raw instructions and execution constraints
from fine-grained descriptions, retaining the strengths of both.
The inverted-U trend suggests that fine-grained language should
augment, not replace, raw task instructions.

\subsection{Architecture and Data-Scale Effects}
\label{sec:arch_scale}

\textbf{FG supervision narrows the architecture gap.}
Comparing StarVLA-OFT and StarVLA-GR00T on the same dataset (RDT), OFT is
clearly stronger under Raw-only supervision (gap of 6.4/6.6 on Easy/Hard), but
the gap shrinks as FG ratio increases and nearly vanishes under FG-only
(0.8/0.5). This suggests that dense language supervision alleviates a
supervision bottleneck, reducing the policy's dependence on decoder
architecture choice.

\textbf{FG supervision benefits more from larger data scale.}
Comparing RDT-OFT and AlohaMix-OFT, the gain from FG supervision is larger
on the bigger AlohaMix dataset. The FG-only improvement over Raw-only grows
from +1.4/+2.0 (RDT) to +6.5/+4.7 (AlohaMix). As trajectory diversity grows,
dense action-aligned language has more distinct execution patterns to bind to.
Together with the architecture result above, this suggests that fine-grained
supervision is not merely an incidental improvement for current-scale training:
it represents a scalable supervision axis beyond a single
architecture or dataset scale.

Detailed per-setting numbers supporting both observations are reported in
Appendix~\ref{app:robotwin_analysis},
Table~\ref{tab:robotwin_analysis_summary}.

\FloatBarrier
\subsection{Fine-Grained Language Enables Factor-Level Steerable Control}
\label{sec:language_critical}

Overall task success can hide instruction violations: a policy
may complete the goal-level task while using the wrong arm,
approaching from the wrong direction, or rotating in the wrong
direction. We therefore examine the five single-factor columns in
Table~\ref{tab:real}, where each column isolates exactly one
language-specified control attribute while holding the visual
scene fixed.

Table~\ref{tab:real} shows that FG\,:\,Raw\,$=$\,1\,:\,1
improves every instruction-sensitive factor over Raw-only.
The largest gains appear on attributes invisible to goal-level
language: Pose (24\,$\to$\,47, +23), Color
(22\,$\to$\,40, +18), and Approach (60\,$\to$\,78, +18).
Rotate improves from 76 to 86 (+10), and Arm from 60 to 64
(+4). The gain magnitude correlates with how much each factor
is underspecified by raw instructions: object pose, color, and
approach direction receive no guidance in goal-level language,
while rotation direction and arm selection are occasionally
implied by task context. These results show that fine-grained
supervision improves not only overall task completion, but also
execution compliance on the specific control attribute
specified by the instruction.

The OOD actor-target probe reveals a different pattern.
The FG\,:\,Raw\,$=$\,1\,:\,2 and 1\,:\,1 settings achieve the
highest OOD scores (5 and 10 respectively), compared with 0 for
Raw-only, suggesting that mixed supervision strengthens
individual factor grounding. However, this does not translate
into full task completion because the policy still fails to bind
the selected arm to the unseen target receptacle. Thus,
\framework{} improves factor-level controllability, but full
compositional generalization remains unsolved.

\subsection{Limitations}
\label{sec:error_limitations}

The remaining real-world failures fall into two categories.
The first is \emph{grounding failure}, where the policy selects
the wrong object, arm, or target despite the language specifying
the correct factor. The second is \emph{execution failure}, where
the correct factor is selected but the physical manipulation
fails, such as unstable grasping, incomplete rotation, or
inaccurate placement. The OOD actor-target probe further shows a
compositional limitation: increasing FG supervision improves
active-arm grounding, but does not reliably solve novel
actor-target binding.

Our framework also has several limitations. \vlmname{} reduces
annotation cost but does not fully remove human verification.
Real-world validation is still limited to a tabletop dual-arm
platform and a small set of targeted steerability tasks. Our
benchmark metrics also have a scope limitation: the caption-track
Anti-Hallucination score is measured only over action steps, because
the ground-truth atomic actions exhaustively cover all executed
actions and thus make fabrication well-defined, whereas the other nine
fine-grained dimensions cannot be exhaustively enumerated---an
unmatched description there may be a correct supplementary detail
rather than a fabrication, so hallucination cannot be reliably defined
for them. Finally,
following fine-grained execution instructions in physical
environments raises safety concerns; future systems should
combine fine-grained language following with feasibility and
safety checks.

%% file: sections/related_work.tex
\section{Related Work}
\label{sec:related}

\textbf{VLA policy learning and sparse trajectory language.}
Recent VLA policies such as RT-2~\citep{rt2}, OpenVLA~\citep{openvla},
$\pi_0$~\citep{pi0}, and Octo~\citep{octo} leverage pretrained
vision-language models and large demonstration datasets such as Open
X-Embodiment~\citep{oxe}, DROID~\citep{droid}, and BridgeData
V2~\citep{bridgedata_v2}. While these efforts substantially improve
generalist policy learning, their paired language supervision remains sparse:
each trajectory is annotated with a goal-level task name that specifies the
desired outcome but omits execution details such as which arm to use, how to
approach the object, or what motion path to follow.

\textbf{Fine-grained supervision for manipulation.}
Several works enrich supervision beyond trajectory-level labels.
Galaxea~\citep{galaxea_g0}, RoboCOIN~\citep{robocoin}, and
RoboInter~\citep{robointer} introduce subtask or hierarchical annotations;
STEER~\citep{steer} and PartInstruct~\citep{partinstruct} study low-level or
part-level instruction following. These annotations are typically organized
around stages, primitives, or object parts rather than full process-level
descriptions. FineVLA instead provides process-level, action-aligned
supervision across a ten-dimensional schema that unifies actor choice, contact
patterns, motion trajectories, state transitions, and recovery behavior---and
uses this supervision consistently for data construction, VLM training,
benchmark evaluation, and policy learning.

\textbf{Robotic video understanding and scalable annotation.}
General video-language models such as Qwen3-VL~\citep{qwen3vl} and
Qwen3.5-Omni~\citep{qwen35omni} provide strong foundations for video
captioning, while embodied benchmarks such as RoboVQA~\citep{robovqa},
RoboBench~\citep{robobench}, and HanDyVQA~\citep{handyvqa} evaluate spatial
reasoning, affordances, and hand-object dynamics. Dense captioning methods
such as Wolf~\citep{wolf}, DIAL~\citep{dial}, and
RoboAnnotatorX~\citep{roboannotatorx} further improve annotation scalability.
However, general captions do not necessarily align with robot action. FineVLA
closes this gap by connecting robotic video understanding directly to VLA
policy learning: \vlmname{} generates action-aligned descriptions,
\benchmarkname{} evaluates execution-level understanding, and \policyname{}
tests whether such supervision improves instruction-sensitive control.

\textbf{Steerable robot foundation models.}
Recent robot foundation models increasingly emphasize
instruction-steerable behavior, where policies should follow not
only task goals but also execution-level
constraints~\citep{pi0_7,lingbot_vla,isaac_gr00t_github,gen1_blog_2026}.
However, the data construction and evaluation infrastructure
behind many such systems remains limited or closed.
\framework{} complements these efforts by providing an open
action-aligned annotation pipeline, a held-out benchmark, a
scalable annotator, and a controlled policy-training study.

%% file: sections/conclusion.tex
\section{Conclusion}
\label{sec:conclusion}

We presented \textbf{\framework{}}, a framework that reframes
steerable VLA learning as an action-instruction alignment
problem: language supervision should specify not only \emph{what}
task to complete, but also the execution-level choices that
determine \emph{how} the robot completes it.

Starting from 972,247 trajectories across 10 open-source
datasets, \toolname{} produces 47,159 human-verified
trajectories with process-level annotations spanning ten
fine-grained dimensions. \vlmname{}, fine-tuned on this data,
serves as a scalable annotator that achieves 68.2\% VQA accuracy
and 82.2\% captioning score on the held-out \benchmarkname{}.
\policyname{}, trained under controlled FG:Raw instruction
mixtures, reaches \textbf{86.8\%}/\textbf{82.5\%} on
AlohaMix-OFT Easy/Hard in RoboTwin simulation, and
\textbf{62.7}/100 in real-world dual-arm manipulation
(vs.\ 49.9 for Raw-only), with the largest per-factor gains on
execution-sensitive attributes such as pose (+23), color (+18),
and approach direction (+18).

Two key findings emerge. First, fine-grained supervision does not
sacrifice goal-level task success; it consistently improves over
raw-only baselines across architectures, data scales, and
environments. Second, fine-grained and raw instructions are
complementary: the inverted-U trend across all settings shows
that the best steerable control comes from mixing both---raw
instructions specify \emph{what} to achieve, while fine-grained
descriptions specify \emph{how} to execute it. Third,
fine-grained supervision directly improves factor-level steerable
control: in real-world evaluation, the largest gains appear on
execution-sensitive attributes such as color, pose, and approach
direction, where goal-level instructions provide no guidance.

We release \toolname{}, \vlmname{}, \benchmarkname{}, and
\policyname{} checkpoints and training code to support
reproducible research on steerable VLA policies. Remaining
challenges include compositional generalization to unseen
instruction combinations, validation across broader embodiments
and task domains, and integrating feasibility and safety checks
for fine-grained language following in physical deployment.

%% file: sections/appendix.tex

This appendix is organized to mirror the logic of the main paper and collects
detailed methodology, extended analyses, and reproducibility information that
support the claims in Sections~\ref{sec:tool}--\ref{sec:analysis}.

\subsection{FineVLA-Tool Details}
\label{sec:app_tool}
\label{app:tool}

This section provides implementation and annotation details for
Section~\ref{sec:tool}. It explains how heterogeneous robot trajectories are
converted into a unified action-aligned supervision corpus and how human review
is used to maintain annotation quality.

\subsubsection{Data Sources and Format Conversion}
\label{sec:app_tool_sources}
\label{app:tool_sources}

This subsection supports the data-construction claim in Section~\ref{sec:tool}
by listing the source datasets and the unified-format conversion step used
before filtering, clustering, and annotation.

Table~\ref{tab:tool_source_datasets} summarizes the ten open-source datasets
used by \toolname{} before clustering and representative sampling. We convert
all trajectories to a unified LeRobot~2.1-style format that standardizes RGB
videos, robot states, action sequences, and task metadata across embodiments.
This conversion is a prerequisite for consistent filtering, action-state
canonicalization, and later cross-dataset annotation.

\begin{table}[htbp]
  \caption{\textbf{Source datasets used by \toolname{}.} The table reports
  the ten datasets retained for fine-grained annotation after selecting the
  final source list used in this paper.}
  \label{tab:tool_source_datasets}
  \centering
  \small
  \begin{tabular}{@{}lrr@{}}
    \toprule
    \textbf{Dataset} & \textbf{\# Trajectories} & \textbf{\# Tasks} \\
    \midrule
    BridgeData-V2 & 53,192 & 19,974 \\
    BC-Z & 39,350 & 104 \\
    RT-1 & 87,212 & 599 \\
    Galaxea & 113,000 & 19,627 \\
    RoboMIND-V1 & 73,909 & 436 \\
    RoboMIND-V2 & 258,786 & 723 \\
    RoboCOIN & 104,857 & 807 \\
    RH20T-RoboInter & 82,894 & 147 \\
    RDT-1B & 6,061 & 296 \\
    DROID-RoboInter & 152,986 & 43,026 \\
    \midrule
    Total & 972,247 & 85,739 \\
    \bottomrule
  \end{tabular}
\end{table}

\subsubsection{Action-State Canonicalization}
\label{sec:app_tool_canonicalization}
\label{app:canonicalization}

This subsection supports the canonicalization step in Section~\ref{sec:tool}.
Its role is to make state and action sequences comparable across datasets that
use different temporal conventions and different kinematic parameterizations.

Across datasets, action and state annotations differ in both temporal reference
and kinematic representation. We organize these differences along two axes:
(i)~temporal convention, distinguishing absolute quantities, deltas relative to
current state, and offsets relative to the first frame; and
(ii)~kinematic convention, distinguishing joint-space signals from
end-effector-space signals with multiple rotation encodings. We standardize EEF
rotations to quaternions in \texttt{xyzw} order.

\begin{table}[htbp]
  \caption{\textbf{Canonicalization tokens and their semantics.} Prefixes
  define temporal reference, while suffixes define the parameterization of
  robot state or action variables.}
  \label{tab:prefix_suffix_semantics}
  \centering
  \small
  \begin{tabular}{@{}lll@{}}
    \toprule
    \textbf{Token} & \textbf{Meaning} & \textbf{Notes} \\
    \midrule
    \texttt{abs} & Absolute value in a global/world frame & Used by all states \\
    \texttt{delta} & Increment relative to the current state & Action only \\
    \texttt{rel} & Offset relative to the first frame & Action only \\
    \midrule
    \texttt{joint} & Joint / gripper / hand coordinates & Non-EEF modalities \\
    \texttt{rotvec} & Rotation vector (axis-angle) & 3D rotation code \\
    \texttt{quat} & Quaternion in \texttt{xyzw} order & Canonical quaternion form \\
    \texttt{wxyz} & Quaternion in \texttt{wxyz} order & Scalar-first order \\
    \texttt{euler} & Euler angles in XYZ order & 3D rotation code \\
    \bottomrule
  \end{tabular}
\end{table}

For non-EEF variables such as joints, grippers, or hand states, the state is
always treated as an absolute quantity. In contrast, the raw action may be
stored as an absolute command or a delta command, and is therefore converted
into one of \texttt{abs\_joint}, \texttt{delta\_joint}, or
\texttt{rel\_joint}. For EEF variables, each pose is represented as 3D
position plus a rotation code; the state remains absolute, while the action may
use any of the three temporal-reference prefixes.

\begin{table}[htbp]
  \caption{\textbf{Canonicalization rules across modality types.} State and
  action variables admit different canonical type sets for non-EEF and EEF
  modalities.}
  \label{tab:modality_type_space}
  \centering
  \small
  \begin{tabular}{@{}p{0.28\linewidth}p{0.25\linewidth}p{0.37\linewidth}@{}}
    \toprule
    \textbf{Field type} & \textbf{State options} & \textbf{Action options} \\
    \midrule
    Non-EEF
    & \texttt{abs\_joint}
    & \texttt{abs\_joint}, \texttt{delta\_joint}, \texttt{rel\_joint} \\
    \addlinespace
    EEF (left/right end effector)
    & \texttt{abs\_rotvec}, \texttt{abs\_quat}, \texttt{abs\_wxyz},
      \texttt{abs\_euler}
    & All 12 combinations formed by
      \{\texttt{abs}, \texttt{delta}, \texttt{rel}\}
      $\times$
      \{\texttt{rotvec}, \texttt{quat}, \texttt{wxyz}, \texttt{euler}\} \\
    \midrule
    Count summary
    & 5 state tags overall
    & 15 action tags overall \\
    \bottomrule
  \end{tabular}
\end{table}

A minimal example is as follows. Let $s_t^{\text{joint}} \in \mathbb{R}^d$
be the current joint state and let the raw action be a delta command
$\Delta a_t^{\text{joint}}$. The absolute next-state target is
\[
\hat{s}_{t+1}^{\text{joint}} = s_t^{\text{joint}} + \Delta a_t^{\text{joint}},
\]
while the relative action with respect to the first frame is
\[
a_{t+1,\mathrm{rel}}^{\text{joint}}
= \hat{s}_{t+1}^{\text{joint}} - s_1^{\text{joint}}.
\]
For EEF trajectories, we first convert the raw orientation code to a canonical
\texttt{xyzw} quaternion, compose delta poses with the current absolute pose if
necessary, and only then derive the desired absolute, delta, or first-frame
relative action target.


\subsubsection{Quality Filtering and DTW Consistency Check}
\label{sec:app_tool_filtering}

This subsection supports the cleaning step in Section~\ref{sec:tool}. It
explains how we remove invalid videos and trajectories whose actions and states
are inconsistent after canonicalization.

We first perform video-level filtering to remove trajectories with invalid or
missing videos, extremely short duration, large black-frame segments, or other
obvious recording failures. We then apply action-state consistency filtering on
canonicalized trajectories. Intuitively, a valid demonstration should exhibit a
state evolution that is compatible with the recorded action sequence after both
have been converted into a common representation. We therefore measure
trajectory-level consistency using DTW and reject samples whose action-state
DTW distance exceeds a dataset-specific threshold. This stage removes corrupted
logs, mismatched control conventions, and trajectories whose recorded actions do
not explain the observed state change.

\subsubsection{Action-Based Clustering and Representative Sampling}
\label{sec:app_tool_clustering}
\label{app:tool_clustering}

This subsection provides full implementation details for the
representative-sampling step described in Section~\ref{sec:tool}. The goal is
to reduce redundancy in large robot corpora while preserving genuinely distinct
execution patterns, so that a fixed annotation budget covers the widest
possible range of manipulation strategies.

\paragraph{Pipeline overview.}
The clustering pipeline proceeds in four stages:
\begin{enumerate}[leftmargin=*,itemsep=1pt]
  \item \textbf{Canonicalization.} All trajectories within a task are converted
    to their canonical action representation (joint-space or EEF-space with
    quaternion rotations) following the procedure in
    Appendix~\ref{app:canonicalization}.
  \item \textbf{Pairwise DTW distance computation.} For each pair of
    trajectories within a task, we compute the DTW distance using a
    representation-specific frame cost function (defined below).
  \item \textbf{Hierarchical clustering.} Agglomerative clustering with average
    linkage is applied to the pairwise distance matrix, and the number of
    clusters is determined automatically via the largest relative gap in merge
    heights.
  \item \textbf{Representative selection.} Two to three high-quality
    trajectories are selected from each cluster based on proximity to the
    cluster medoid and trajectory quality metrics (video integrity, action
    smoothness).
\end{enumerate}

\paragraph{DTW formulation.}
Open robot datasets are highly redundant: many demonstrations differ only in
speed, minor spatial offsets, or camera viewpoint, while expressing the same
underlying action pattern. Dynamic Time Warping (DTW) handles temporal
misalignment by finding the optimal warping path between two action sequences.
Given two sequences $\mathbf{x}_{1:T}$ and $\mathbf{y}_{1:U}$, DTW minimizes
the cumulative frame-level distance according to
\[
D_{\mathrm{DTW}}(i,j)
= c(\mathbf{x}_i,\mathbf{y}_j)
+ \min\!\bigl\{
    D_{\mathrm{DTW}}(i-1,j-1),\;
    D_{\mathrm{DTW}}(i-1,j),\;
    D_{\mathrm{DTW}}(i,j-1)
  \bigr\}.
\]

\paragraph{Frame cost function.}
The frame cost $c(\cdot,\cdot)$ depends on the action-space representation:

\begin{itemize}[leftmargin=*,itemsep=2pt]
  \item \textbf{Joint-space} ($\texttt{rot\_type} = \texttt{none}$): All joint
    values are min-max normalized to $[0,1]$ per dimension across the task
    group. The frame cost is
    \[
      c_{\text{joint}}(\mathbf{x},\mathbf{y})
      = w_{\text{pos}} \cdot \|\mathbf{j}_x - \mathbf{j}_y\|_2
      + w_{\text{grip}} \cdot |g_x - g_y|,
    \]
    where $\mathbf{j}$ denotes the normalized joint vector and $g$ the gripper
    state.
  \item \textbf{EEF-space} (quaternion or Euler): The frame cost combines
    position, rotation, and gripper terms:
    \[
      c_{\text{eef}}(\mathbf{x},\mathbf{y})
      = w_{\text{pos}} \cdot \|\mathbf{p}_x - \mathbf{p}_y\|_2
      + w_{\text{rot}} \cdot d_{\text{geo}}(\mathbf{q}_x, \mathbf{q}_y)
      + w_{\text{grip}} \cdot |g_x - g_y|,
    \]
    where $\mathbf{p}$ is the 3D position, $d_{\text{geo}}(\mathbf{q}_x,
    \mathbf{q}_y) = 2\arccos(|\mathbf{q}_x \cdot \mathbf{q}_y|)$ is the
    quaternion geodesic distance (handling $\mathbf{q} \equiv -\mathbf{q}$),
    and $g$ is the gripper state. For datasets using Euler angles, orientations
    are first converted to quaternions.
\end{itemize}

Default weights are $w_{\text{pos}} = 1.0$, $w_{\text{rot}} = 1.0$,
$w_{\text{grip}} = 100.0$. The high gripper weight ensures that gripper
open/close transitions---which are critical for distinguishing manipulation
strategies---are not overwhelmed by continuous motion differences.

\paragraph{Pairwise distance computation.}
Since each task typically contains 100--200 trajectories, computing the full
$N \times N$ pairwise DTW distance matrix is tractable (${\sim}$5k--20k pairs
per task). Each DTW distance is normalized by the warping path length to
account for differences in trajectory duration. Computation is parallelized
across CPU cores.

\paragraph{Hierarchical clustering and representative selection.}
We apply agglomerative hierarchical clustering (average linkage) on the DTW
distance matrix. The number of clusters $k$ is selected automatically by
identifying the largest relative gap in the dendrogram merge heights. We then
select two to three representative trajectories per cluster according to
cluster size and trajectory quality (proximity to the cluster medoid).
Applying this procedure to 972,247 raw demonstrations yields 47,159
representative trajectories for fine-grained annotation. This greatly reduces
annotation cost while preserving diversity in manipulation strategy, object
interaction, and motion pattern.

\begin{figure*}[htbp]
  \centering
  \includegraphics[width=\linewidth]{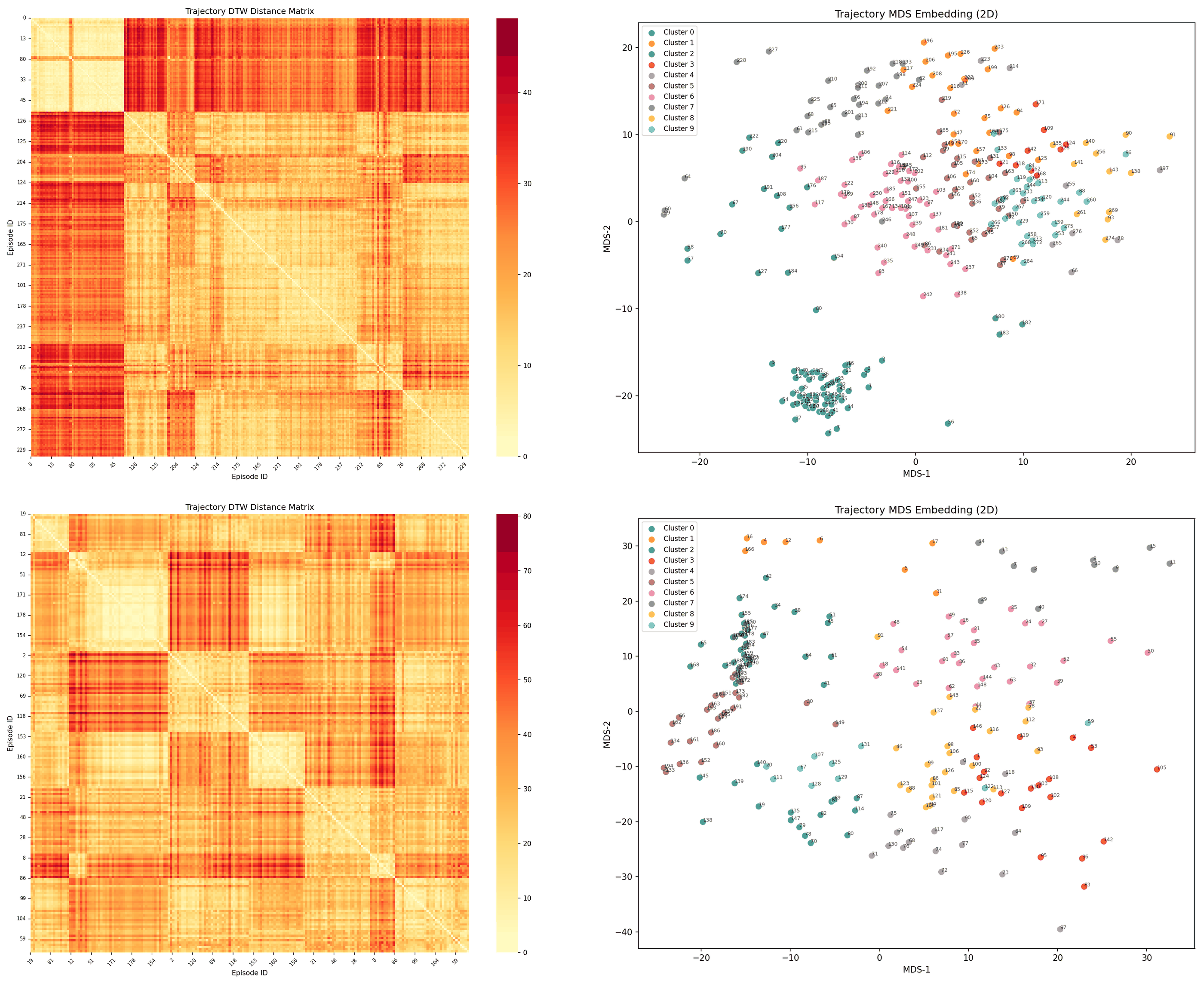}
  \caption{\textbf{Qualitative examples of DTW-based trajectory clustering in
  \toolname{}.} For each task, the left panel shows the pairwise DTW distance
  matrix and the right panel shows a 2D MDS embedding of the same distances.
  Clear cluster structure indicates that trajectories with similar
  manipulation dynamics are grouped together, while differences in gripper
  timing and end-effector motion patterns are separated into different
  clusters. This allows one fine-grained instruction to cover multiple
  redundant demonstrations with the same execution pattern, substantially
  reducing annotation cost while improving data quality.}
  \label{fig:tool_clustering_examples}
\end{figure*}

Figure~\ref{fig:tool_clustering_examples} shows two task-level clustering
examples. In both cases, the DTW distance matrix exhibits clear block
structure, and the corresponding MDS embedding separates execution modes that
differ in gripper timing, contact duration, or end-effector path. This is the
key reason clustering improves annotation efficiency: a single fine-grained
instruction can cover multiple redundant demonstrations that share the same
execution pattern.

\subsubsection{Fine-Grained Annotation Schema}
\label{sec:app_tool_schema}
\label{app:fine_grained_schema}

This subsection supports the annotation schema referenced throughout the main
paper. It defines the ten dimensions used both for data annotation and for
benchmark construction.

{\small
\setlength{\tabcolsep}{3pt}
\begin{table}[p]
\caption{\textbf{Fine-grained annotation schema used by \toolname{}.}
Each representative trajectory in \datasetname{} is annotated along ten
control-relevant dimensions. The same schema is also used by
\benchmarkname{} to construct VQA questions and caption-level atomic
facts.}
\label{tab:fine_grained_schema}
\centering
\begin{tabularx}{\linewidth}{@{}>{\raggedright\arraybackslash}p{1.35cm}XXX@{}}
\toprule
\textbf{Dimension} & \textbf{What it captures} & \textbf{Typical slots}
  & \textbf{Examples} \\
\midrule

\textbf{Action sequence}
&
The ordered manipulation primitives and gripper states that define the
execution process. Captures which actions occur and their temporal order.
&
Primitive action; gripper state; step order; action transition.
&
Open gripper $\rightarrow$ approach mug $\rightarrow$ grasp handle
$\rightarrow$ lift $\rightarrow$ place on tray $\rightarrow$ release.
\\
\midrule

\textbf{Active actor}
&
The robot component or body part that performs the action. Important for
bimanual manipulation and multi-effector platforms.
&
Actor identity; left/right hand; both hands; gripper; finger; arm;
end-effector.
&
The right gripper picks up the cup. Both hands lift the container
together.
\\
\midrule

\textbf{Target object}
&
The object directly manipulated, with visual or spatial attributes to
disambiguate among similar objects.
&
Object category; color; material; shape; size; spatial reference.
&
The red cup on the left. The small metal spoon inside the bowl.
\\
\midrule

\textbf{Initial config.}
&
The pose, state, or spatial relation of the target object before the
action begins.
&
Initial pose; workspace location; attachment state; relational state.
&
The bottle is standing upright on the table. The drawer is partially
open.
\\
\midrule

\textbf{Final config.}
&
The pose, state, or spatial relation of the target object after the
action is completed.
&
Final pose; final location; relational state; object state.
&
The cup ends up on the tray. The drawer is fully closed.
\\
\midrule

\textbf{Contact \& approach}
&
How the robot approaches the target and where physical contact occurs.
Captures control-relevant interaction geometry.
&
Contact region; approach direction; grasp type; contact modality.
&
Grasp the mug by the handle from above. Push the drawer from its front
panel.
\\
\midrule

\textbf{Trajectory \& orientation}
&
The translational path, motion direction, path shape, and rotational
behavior of the object or end-effector.
&
Translation direction; path shape; motion extent; rotation direction;
rotation angle.
&
Lift upward. Move to the right along a straight path. Rotate clockwise
${\sim}90^{\circ}$.
\\
\midrule

\textbf{Object interaction}
&
Incidental or relational effects on objects other than the target.
Captures secondary contacts, collisions, or induced movements.
&
Interaction type; affected object; contact or collision; relational
change.
&
Placing the book causes the stack to shift. The block brushes against
its neighbor.
\\
\midrule

\textbf{Failure \& recovery}
&
Visible execution failures and recovery behaviors, including retries,
slippage, and fail-then-succeed patterns.
&
Failure type; recovery action; retry behavior; attempt count; slippage.
&
The robot first fails to grasp, then retries successfully. The object
slips before being regrasped.
\\
\midrule

\textbf{Body motion}
&
Motion of the robot base, torso, camera, or other non-end-effector body
parts that affects execution.
&
Base translation; base rotation; torso motion; camera motion.
&
The mobile base moves forward before the arm reaches the object.
\\

\bottomrule
\end{tabularx}
\end{table}
}

\subsubsection{Human-in-the-Loop Verification}
\label{sec:app_tool_verification}

This subsection supports the claim that \datasetname{} is human verified. It
summarizes the dimensions checked during review after the initial annotation or
model-assisted draft is produced.

Human reviewers compare each step-level description against the corresponding
video and verify both semantic correctness and temporal alignment. The review
process focuses on the factors most critical for downstream control.

\begin{table}[htbp]
  \caption{\textbf{Verification dimensions in the human review stage.} These
  checks ensure that fine-grained instructions remain temporally aligned and do
  not introduce hallucinated events.}
  \label{tab:app_tool_verification}
  \centering
  \small
  \begin{tabular}{@{}p{0.30\linewidth}p{0.60\linewidth}@{}}
    \toprule
    \textbf{Verification dimension} & \textbf{What is checked} \\
    \midrule
    Temporal order & Whether step order matches the observed execution timeline \\
    Object identity & Whether the manipulated object is correctly identified and disambiguated \\
    Actor identity & Whether the responsible arm / gripper / body part is correctly assigned \\
    Contact region & Whether grasp/contact location and approach side are faithful to the video \\
    Motion direction & Whether translational and rotational directions are described correctly \\
    State transition & Whether initial and final states match the observed change \\
    Hallucinated events & Whether the instruction introduces unsupported actions or recoveries \\
    \bottomrule
  \end{tabular}
\end{table}

\subsection{RoboFine-VLM Details}
\label{sec:app_vlm}
\label{app:vlm}

This section provides supporting details for the RoboFine-VLM component in
Section~\ref{sec:roboactvlm}. It explains the training data format, the known
model setup, and how the resulting model is used as a scalable annotator.

\subsubsection{SFT Data Format}
\label{sec:app_vlm_format}

This subsection supports the supervised fine-tuning setup described in
Section~\ref{sec:roboactvlm}. Each training sample pairs a robot manipulation
video with a prompt asking for a temporally ordered fine-grained action
description.

\begin{table}[htbp]
  \caption{\textbf{Supervised fine-tuning sample format for RoboFine-VLM.}
  Each sample maps a robot video to a step-level, action-aligned description
  derived from \toolname{} annotations.}
  \label{tab:app_vlm_sft_format}
  \centering
  \small
  \begin{tabular}{@{}p{0.24\linewidth}p{0.68\linewidth}@{}}
    \toprule
    \textbf{Field} & \textbf{Description} \\
    \midrule
    Input video & Multi-frame robot manipulation video sampled from a trajectory \\
    Prompt & Instruction asking the model to describe the manipulation process step by step \\
    Output target & Temporally ordered fine-grained action description aligned with \toolname{} annotations \\
    Supervision source & Human-verified video-instruction pairs from \datasetname{} \\
    \bottomrule
  \end{tabular}
\end{table}

\subsubsection{Model and Training Details}
\label{sec:app_vlm_training}

This subsection supports the model description in Section~\ref{sec:roboactvlm}.

\begin{table}[htbp]
  \caption{\textbf{RoboFine-VLM training configuration.}}
  \label{tab:app_vlm_training}
  \centering
  \small
  \begin{tabular}{@{}p{0.33\linewidth}p{0.55\linewidth}@{}}
    \toprule
    \textbf{Item} & \textbf{Value} \\
    \midrule
    Base model & Qwen3.5-397B-A17B \\
    Fine-tuning strategy & Full parameter fine-tuning \\
    Training objective & Supervised fine-tuning for step-level robotic action understanding \\
    Training corpus & FineVLA-Tool-generated and human-verified video-instruction pairs \\
    Training sample count & 47,159 representative trajectories / video-instruction pairs \\
    Output format & Ordered fine-grained action description \\
    \midrule
    GPUs & 256$\times$H200 \\
    Global batch size & 512 \\
    Epochs & 9 \\
    Training steps & 903 \\
    Optimizer & AdamW \\
    Learning rate & 7e-6, decayed to 7e-7 \\
    LR schedule & Linear decay \\
    Wall-clock training time & $\sim$40 hours (1 day 16 hours) \\
    Memory per GPU & $\sim$105\,GB \\
    \bottomrule
  \end{tabular}
\end{table}

\subsubsection{Video Sampling Configuration}
\label{sec:app_vlm_video_sampling}

This subsection details how raw robot videos are sampled into frame sequences
for SFT. We use two configurations depending on the number of camera views
available in each trajectory.

\begin{table}[htbp]
  \caption{\textbf{Video sampling configuration for RoboFine-VLM SFT.}
  Multi-view trajectories concatenate frames from all available cameras,
  while single-view trajectories allow longer temporal context per video.}
  \label{tab:app_vlm_video_sampling}
  \centering
  \small
  \begin{tabular}{@{}p{0.30\linewidth}p{0.28\linewidth}p{0.28\linewidth}@{}}
    \toprule
    \textbf{Parameter} & \textbf{Multi-view} & \textbf{Single-view} \\
    \midrule
    Frame sampling rate & 4 fps & 4 fps \\
    Max frames per video & 512 & 1024 \\
    Max tokens per video & 76k & 224k \\
    \bottomrule
  \end{tabular}
\end{table}

\subsubsection{SFT Prompt Templates}
\label{sec:app_vlm_prompts}

This subsection provides the system prompts used during supervised fine-tuning
of \vlmname{}. We use two prompt variants depending on the number of camera
views.

\paragraph{Single-view prompt.}
The following prompt is used for trajectories with a single camera view:

\begin{framed}
{\small\ttfamily\noindent
You are a robot manipulation video annotator.

\medskip
Task: Decompose the video into an ordered sequence of fine-grained steps. Each step must describe exactly one visible physical movement by the robot.

\medskip
For each step, include the following when visually confirmed:\\
-- action and gripper state (e.g., grasp, pick up, place, push, rotate; open \textrightarrow{} close, release, maintain grasp)\\
-- active actor (left hand, right hand, both hands, gripper, finger)\\
-- target object using the task instruction name; disambiguate similar objects by position or appearance (e.g., the red cup on the left, the front-most block)\\
-- contact region and approach direction (e.g., by the handle from above, at the top edge from the right)\\
-- object initial state or location if relevant (e.g., upright, lying flat, inside the container)\\
-- trajectory and orientation change (e.g., move forward, lift up, rotate clockwise/counter-clockwise)\\
-- final placement location and final pose if the object is repositioned\\
-- interaction with other objects (e.g., collision, dragging, tipping, displacement)\\
-- failures, retries, slippage, or fail-then-succeed patterns\\
-- body motion such as base, torso, or camera movement

\medskip
Rules:\\
-- One step = one physical movement.\\
-- Describe only visible facts. Do not infer hidden actions, intent, or occluded contact.\\
-- Use the robot's egocentric frame: left, right, forward, backward, up, down.\\
-- Keep descriptions short, action-oriented, and literal.\\
-- If a detail is unclear or not visible, omit it.\\
-- Mention dual-arm coordination only when both arms are active; label it as stabilize-and-act, simultaneous, sequential, or handoff.

\medskip
Output JSON only:\\
\{\quad"Step1": "...", "Step2": "...", "StepN": "..." \quad\}
}
\end{framed}

\paragraph{Multi-view prompt.}
The following prompt is used for trajectories with three synchronized camera
views (one main view and two wrist views):

\begin{framed}
{\small\ttfamily\noindent
You are a robot manipulation video annotator.

\medskip
Task: Analyze three synchronized views and decompose the robot's behavior into an ordered sequence of atomic physical movements.

\medskip
View use:\\
-- Main View: primary view for all annotation decisions, including step boundaries, action order, active arm, global motion, object relations, and final results.\\
-- Wrist Views (left/right): auxiliary only. Use them only to refine local details such as gripper state, contact region, approach direction, or slight slip.\\
-- If Wrist Views conflict with the Main View, follow the Main View.

\medskip
Step requirements:\\
For each step, include the following when visually confirmed:\\
-- action and gripper state (e.g., grasp, pick up, place, push, rotate; open \textrightarrow{} close, release, maintain grasp)\\
-- active actor (left hand, right hand, both hands, gripper, finger)\\
-- target object using the task instruction name; disambiguate similar objects by position or appearance (e.g., the red cup on the left, the front-most block)\\
-- contact region and approach direction (e.g., by the handle from above, at the top edge from the right)\\
-- object initial state or location if relevant (e.g., upright, lying flat, inside the container)\\
-- trajectory and orientation change (e.g., move forward, lift up, rotate clockwise/counter-clockwise)\\
-- final placement location and final pose if the object is repositioned\\
-- interaction with other objects (e.g., collision, dragging, tipping, displacement)\\
-- failures, retries, slippage, or fail-then-succeed patterns\\
-- body motion such as base, torso, or camera movement

\medskip
Rules:\\
-- One step = one physical movement.\\
-- Start a new step when the robot changes primitive action, target, or contact state.\\
-- Describe only visible facts. Do not infer hidden actions, intent, or occluded contact.\\
-- If a detail is unclear or not visually confirmed, omit it.\\
-- Use the robot's egocentric frame: left, right, forward, backward, up, down.\\
-- Keep descriptions concise, action-oriented, and literal.\\
-- Mention dual-arm coordination only when both arms actively contribute to the same manipulation event; use stabilize-and-act, simultaneous, or handoff when applicable.

\medskip
Output JSON only:\\
\{\quad"Step1": "...", "Step2": "...", "StepN": "..." \quad\}
}
\end{framed}

\subsubsection{Inference and Scalable Annotation}
\label{sec:app_vlm_inference}

This subsection supports the scalable-annotation claim in
Section~\ref{sec:roboactvlm}. After fine-tuning, \vlmname{} serves both as an
evaluation model on \benchmarkname{} and as a scalable annotator for new robot
trajectories. Given a video, the model generates a temporally ordered action
description in the same semantic space as the \toolname{} annotations. These
auto-generated descriptions can then be post-processed and human-verified,
allowing fine-grained supervision to be extended beyond the manually reviewed
subset without changing the schema.


\subsection{RoboFine-Bench Details}
\label{sec:app_bench}
\label{app:benchmark}

This section provides supporting details for \benchmarkname{}, the benchmark
component introduced in Section~\ref{sec:roboactbench} and analyzed in
Section~\ref{sec:benchmark_results}. It covers benchmark construction, the VQA
and caption tracks, evaluation robustness, and additional result analysis.

\subsubsection{Benchmark Construction and Statistics}
\label{sec:app_bench_stats}

This subsection supports the benchmark-construction claim in
Section~\ref{sec:roboactbench}. It summarizes the scale, sampling strategy,
and granularity of the benchmark.

\paragraph{Sampling strategy.}
We sample exactly 50 trajectories from each of the 10 source datasets,
yielding 500 benchmark videos in total. This uniform allocation ensures that
no single dataset dominates the benchmark and that all 10 data sources
contribute equally to the evaluation. The 500 benchmark trajectories are
used exclusively for evaluation; \benchmarkname{} contains no training split,
and all benchmark videos are disjoint from the \vlmname{} SFT training set
(see Section~\ref{sec:roboactbench}).

\begin{table}[htbp]
  \caption{\textbf{Summary statistics of \benchmarkname{}.} The benchmark is
  built to test fine-grained robotic action understanding rather than only
  task-level recognition.}
  \label{tab:app_benchmark_stats}
  \centering
  \small
  \begin{tabular}{@{}lr@{}}
    \toprule
    \textbf{Statistic} & \textbf{Value} \\
    \midrule
    Videos & 500 (50 per dataset $\times$ 10 datasets) \\
    Source datasets & 10 \\
    Robot embodiments & 32 \\
    Atomic facts & 11,631 \\
    Avg. annotated steps per sample & 4.3 \\
    Avg. atomic facts per sample & 23.3 \\
    VQA questions & 1,030 \\
    \bottomrule
  \end{tabular}
\end{table}

\subsubsection{VQA Track Details}
\label{sec:app_bench_vqa}
\label{app:vqa_dim_mapping}

This subsection supports the VQA track of Section~\ref{sec:roboactbench}. It
explains how questions are constructed, how ten annotation dimensions are
aggregated into three reporting axes, and how answers are scored.

We construct 1,030 questions that probe execution-level manipulation details
across the same ten dimensions used in the annotation schema. The question
counts per dimension are given in Table~\ref{tab:vqa_question_stats}. For
reporting, these dimensions are aggregated into three higher-level axes:
Entity and Scene Grounding, Action and Motion Understanding, and Interaction
and State Reasoning.

\begin{table}[htbp]
  \caption{\textbf{VQA question distribution across capability dimensions.}
  This table supports the scale and coverage of the VQA track.}
  \label{tab:vqa_question_stats}
  \centering
  \small
  \begin{tabular}{@{} l c @{}}
    \toprule
    \textbf{Dimension} & \textbf{\# Questions} \\
    \midrule
    Action sequence & 218 \\
    Active actor & 52 \\
    Body motion & 10 \\
    Contact and approach & 155 \\
    Failure and recovery & 14 \\
    Final configuration & 182 \\
    Initial configuration & 77 \\
    Object interaction & 26 \\
    Target object & 85 \\
    Trajectory and orientation & 211 \\
    \midrule
    Total & 1030 \\
    \bottomrule
  \end{tabular}
\end{table}

\begin{table}[htbp]
  \caption{\textbf{VQA dimension mapping.} The ten fine-grained annotation
  dimensions are grouped into three reporting axes for \benchmarkname{} VQA
  evaluation.}
  \label{tab:vqa_dim_mapping}
  \centering
  \small
  \begin{tabular}{@{} l l p{0.52\linewidth} @{}}
    \toprule
    \textbf{Axis} & \textbf{Abbr.} & \textbf{Constituent dimensions} \\
    \midrule
    Entity \& Scene Grounding & Gnd.
      & Active actor, Target object, Initial configuration \\
    \addlinespace
    Action \& Motion Understanding & Act.
      & Action sequence, Contact \& approach,
        Trajectory \& orientation, Body motion \\
    \addlinespace
    Interaction \& State Reasoning & State
      & Object interaction, Final configuration,
        Failure \& recovery \\
    \midrule
    Overall & ---
      & Average over all ten dimensions \\
    \bottomrule
  \end{tabular}
\end{table}

Representative VQA prompts include action recognition, temporal ordering,
object grounding, and state-change questions. Examples are shown below:

\begin{itemize}[leftmargin=*,itemsep=1pt]
\item \textbf{Action Recognition}: ``What is the robot doing?
  (A)~Grasping (B)~Pushing (C)~Lifting (D)~Placing''
\item \textbf{Temporal Ordering}: ``Which happens first? (A)~Gripper
  opens (B)~Arm moves to cup (C)~Cup lifted (D)~Cup placed''
\item \textbf{Object Interaction}: ``Which object is the right gripper
  interacting with? (A)~Red block (B)~Blue cup (C)~Green plate
  (D)~None''
\item \textbf{State Change}: ``State of the drawer after action?
  (A)~Fully open (B)~Half open (C)~Closed (D)~Removed''
\end{itemize}

Answer scoring is deterministic. Multiple-choice questions are evaluated by
option matching; yes/no questions are evaluated by normalized string
comparison; and numeric questions are evaluated by value extraction.

\paragraph{VQA question generation prompt.}
The following prompt is used to generate VQA questions from human-reviewed
ground-truth annotations. It operates in two modes: \emph{conflict-based}
(when model-generated steps disagree with GT) and \emph{GT-only} (when the two
are aligned), ensuring questions target both common errors and diverse
fine-grained facts.

\begin{framed}
{\small\ttfamily\noindent
You are a fine-grained robotics video QA-set builder.

\medskip
\textbf{Input:} A JSON array. Each sample contains: sample\_id,
fineGrainedSteps (model-generated, may contain errors), GT (human-reviewed,
always trustworthy).

\medskip
\textbf{Task Overview}

\medskip
Two modes of question generation:

\medskip
\textbf{Mode A --- Conflict-based QA} (when GT and fineGrainedSteps conflict):
Generate questions targeting specific disagreements. The answer must always
come from GT. Facts that GT describes but fineGrainedSteps omits: you MAY ask.
Facts that fineGrainedSteps describes but GT omits: do NOT ask.

\medskip
\textbf{Mode B --- GT-only QA} (when GT and fineGrainedSteps are similar):
Generate questions purely from GT content. Randomly select 3--5 dimensions from
the 13 dimensions below. Do NOT always pick the same dimensions.

\medskip
For every sample, generate \textbf{3--5 QA pairs}. At most 3 GT-only (Mode B)
questions per sample.

\medskip
\textbf{13 Capability Dimensions:}
\begin{enumerate}[leftmargin=*,itemsep=0pt,parsep=0pt]
\item action\_primitive --- fundamental action type (grasp, push, rotate, etc.)
\item actor\_identity --- which arm/hand/gripper performs the action
\item object\_recognition --- object category, color, material, shape, size
\item object\_disambiguation --- distinguishing similar objects via spatial/attribute cues
\item contact\_region --- specific part where gripper contacts the object
\item source\_state\_or\_location --- initial state/position before manipulation
\item trajectory\_and\_orientation --- direction, path, or rotation during motion
\item placement\_specification --- final target location or spatial relation
\item interaction\_with\_other\_objects --- contact/disturbance of non-target objects
\item success\_failure\_retry --- whether the action succeeds, fails, or retries
\item gripper\_state --- open/close/release state at a specific moment
\item temporal\_order\_and\_step\_boundary --- ordering of steps and boundaries
\item body\_motion --- robot base/torso/camera movement
\end{enumerate}

\medskip
\textbf{Dimension Balancing:} For Mode B, randomly select dimensions per sample.
Do NOT ask two questions on the same dimension within one sample. Across the
batch, aim for roughly equal coverage.

\medskip
\textbf{Answer Types:}\\
--- multiple\_choice: 4--8 mutually exclusive options, no ``all/none of the above''\\
--- yes\_no: answer exactly ``yes'' or ``no''\\
--- number: concise Arabic numeral

\medskip
\textbf{Question Writing Rules:}\\
--- Each question tests exactly ONE atomic fact\\
--- Must be answerable by watching the video\\
--- Do NOT ask broad questions (e.g., ``What does the robot do?'')\\
--- Do NOT use visually similar colors as distractors

\medskip
\textbf{Output:} Valid JSON array only. Each item contains sample\_id, status,
qas (3--5 items with question\_id, mode, capability, answer\_type, question,
options, answer, reference\_text).
}
\end{framed}

\subsubsection{Caption Track Details}
\label{sec:app_bench_caption}

This subsection supports the caption track of Section~\ref{sec:roboactbench}.
It defines the two evaluation settings, the atomic-fact representation, and the
metrics used to score generated captions.

In the \emph{easy} setting, the original task instruction is provided to the
model. In the \emph{hard} setting, the model must infer the manipulation
process from visual observations alone. Human-reviewed annotations are
converted into atomic facts, and each generated caption is aligned against this
fact set using the labels \emph{match}, \emph{partial match},
\emph{contradiction}, \emph{omission}, and \emph{hallucination}. We report
three aspect-specific metrics computed from the alignment counts.
Let $M$, $P$, $C$, and $O$ denote the number of GT facts labeled as match,
partial, contradiction, and omission, respectively.
Define $A = M + P + C$ (the number of GT facts addressed by the caption) and
$G = M + P + C + O$ (total GT facts).
Let $H$ be the number of hallucinated action events and $S$ be the total
number of action steps in the generated caption. The metrics are:
\begin{align}
\text{Consistency} &= \frac{M + 0.5\,P}{A}, \\[4pt]
\text{Coverage} &= \frac{M + 0.5\,P}{G}, \\[4pt]
\text{Anti-Hallucination} &= 1 - \frac{H}{S}, \\[4pt]
\text{Overall} &= \frac{\text{Consistency} + \text{Coverage} + \text{Anti-Hallucination}}{3}.
\end{align}

\paragraph{Scope of the Anti-Hallucination metric.}
Anti-Hallucination is deliberately computed only over the
\textit{action\_sequence} dimension: $H$ counts fabricated
\emph{action events} and $S$ is the number of action steps in the
generated caption. We restrict hallucination scoring to action steps
because this is the only dimension for which fabrication can be defined
unambiguously. The ground-truth atomic actions are guaranteed to
enumerate \emph{all} actions executed in the trajectory, so any action
event in the caption without a corresponding ground-truth action is
genuinely fabricated. For the other nine dimensions this guarantee does
not hold: the ground-truth atomic facts do not exhaustively enumerate
every valid descriptive statement (e.g., appearance, spatial relations,
or object-state details), so a caption statement absent from the
ground-truth fact set may be a \emph{correct supplementary observation}
rather than a hallucination. Because these two cases cannot be reliably
separated for the non-action dimensions, we report Anti-Hallucination
only over action steps and treat this restricted scope as a limitation
of the metric (Section~\ref{sec:error_limitations}).

\begin{table}[htbp]
  \caption{\textbf{Caption-track metric and alignment-label definitions.}
  This table clarifies how atomic-fact alignment is converted into the reported
  caption metrics.}
  \label{tab:app_caption_metrics}
  \centering
  \small
  \begin{tabular}{@{}p{0.24\linewidth}p{0.64\linewidth}@{}}
    \toprule
    \textbf{Item} & \textbf{Definition} \\
    \midrule
    Consistency & Whether the facts stated in the generated caption are correct with respect to the ground truth \\
    Coverage & How much of the ground-truth fact set is recovered by the generated caption \\
    Anti-Hallucination & Whether the caption avoids fabricated actions or states unsupported by the ground truth \\
    Match & Generated fact is correct and aligned to a ground-truth fact \\
    Partial match & Generated fact overlaps with a ground-truth fact but is incomplete or slightly imprecise \\
    Contradiction & Generated fact conflicts with the ground truth \\
    Omission & Ground-truth fact is missing from the generated caption \\
    Hallucination & Generated fact has no support in the ground truth \\
    \bottomrule
  \end{tabular}
\end{table}

\subsubsection{Prompt Templates and Evaluation Protocol}
\label{sec:app_bench_prompts}

This subsection provides the exact prompts and evaluation pipelines used for
\benchmarkname{}. All evaluated models (Qwen3-VL-Plus, Qwen3.5-Plus,
Doubao-Seed-2.0-Pro, Gemini-3.1-Pro, GPT-5.4, and \vlmname{}) receive the
same visual input format, prompting protocol, and sampling budget within each
track.

\paragraph{VQA evaluation pipeline.}
For each benchmark sample, multi-view video frames are extracted and labeled
(e.g., \texttt{[View: head\_rgb]}, \texttt{[View: left\_wrist\_rgb]}). All
questions for one sample are batched into a single prompt. The model is
instructed to return structured JSON answers. Multiple-choice options are
randomly shuffled per question (seeded by question ID) to prevent positional
bias. Answers are scored deterministically: yes/no questions by normalized
string comparison, number questions by value extraction, and multiple-choice
questions by option-letter matching.

The VQA system prompt is:

\begin{framed}
{\small\ttfamily\noindent
You are an expert robot manipulation video analyst. You will be shown frames
from a robot manipulation video with one or more camera views. Each view is
labeled (e.g., [View: head\_rgb], [View: left\_wrist\_rgb]). Use information
from ALL views to answer the questions. Answer based ONLY on what you observe
in the video frames. Be precise and concise.
}
\end{framed}

The VQA user prompt template is:

\begin{framed}
{\small\ttfamily\noindent
Watch the video frames carefully and answer ALL of the following questions
about this robot manipulation video.

\medskip
\{questions\_block\}

\medskip
You MUST return your answers as a JSON object. Use the FULL question\_id shown
in brackets [] as the key (NOT the short Q1/Q2 label):

\medskip
\{"answers": \{"<full\_question\_id>": "<your\_answer>", ...\}\}

\medskip
Rules:\\
-- For yes\_no questions: answer EXACTLY "yes" or "no"\\
-- For number questions: answer with JUST a number (e.g. "3")\\
-- For multiple\_choice questions: answer with JUST the option letter (e.g. "B")\\
-- Do NOT include explanations in the JSON values --- only the short answer\\
-- You MUST answer ALL questions
}
\end{framed}

\paragraph{Caption evaluation pipeline.}
The caption track proceeds in two stages:

\textbf{Stage~1: Caption generation.} Given video frames and a prompt asking
for temporally ordered step-level action descriptions, each evaluated model
generates a manipulation caption. In the easy setting, the original task
instruction is provided; in the hard setting, only video frames are given.

\textbf{Stage~2: LLM-based alignment judging.} The generated caption and the
pre-extracted ground-truth atomic facts (grouped by capability dimension) are
passed to an LLM judge (GPT-5.4-Pro by default). The judge evaluates each GT
atomic fact against the raw caption and assigns one of four labels:
\emph{match} (caption correctly states the fact), \emph{partial} (caption
addresses the event but is materially coarser or incomplete),
\emph{contradiction} (caption gives a conflicting value), or \emph{omission}
(caption does not mention the fact). Additionally, the judge identifies
\emph{hallucinated actions}: action events in the caption that have no
correspondence in any GT action\_sequence fact.

The alignment judge prompt specifies detailed semantic tolerance policies
(color-family equivalence, synonym matching, compatible spatial wording, actor
naming equivalence) and strict rules for hallucination detection
(only genuinely fabricated actions with no GT basis are flagged). The judge
returns a structured JSON with per-fact labels, caption evidence, and aggregate
counts. The three reported metrics are then computed from these counts:
\begin{itemize}[leftmargin=*,itemsep=1pt]
  \item \textbf{Consistency}: fraction of caption-addressed GT facts that are
    match or partial (not contradiction).
  \item \textbf{Coverage}: fraction of GT facts labeled match or partial
    (not omission).
  \item \textbf{Anti-Hallucination}: penalizes fabricated action events not
    supported by GT.
\end{itemize}

The full alignment judge system prompt is provided below:

\begin{framed}
{\small\ttfamily\noindent
You are an expert evaluator for fine-grained robot manipulation captions.

\medskip
You receive:\\
1. Pre-extracted GT atomic facts (structured, grouped by capability dimension).\\
2. A raw AI-generated caption (a list of step descriptions, NOT pre-extracted
into atomic facts).

\medskip
Your task is to evaluate each GT atomic fact against the raw caption text and
determine:\\
-- For each GT fact: is it match, partial, contradiction, or omission?\\
-- Additionally, identify any hallucinated action events in the caption that
do NOT appear in the GT action\_sequence facts.

\medskip
GT Fact Evaluation Rules:\\
-- match: caption clearly states or implies the same information as the GT fact.\\
-- partial: caption addresses the same event but is materially coarser or
incomplete.\\
-- contradiction: caption addresses the same event but gives a conflicting
value.\\
-- omission: caption does not address this GT fact at all.

\medskip
Hallucination Detection (action\_sequence only):\\
A hallucinated action must describe a distinct, meaningful action event that no
GT action\_sequence fact covers, is not a sub-action of a matched GT action,
and is not a gripper state change accompanying a matched action.

\medskip
Output: structured JSON with per-fact labels, caption evidence, and summary
counts (match + partial + contradiction + omission == total\_gt\_facts).
}
\end{framed}

\subsubsection{Judge Robustness}
\label{sec:app_bench_robustness}
\label{app:caption_judge_robustness}

This subsection supports the judge-robustness claim made in
Section~\ref{sec:benchmark_results}. It compares caption scores obtained with
GPT-5.4-Pro and Gemini-3.1-Pro as alignment judges.

The main-paper caption results use GPT-5.4-Pro as the alignment judge. To test
whether the benchmark conclusions are sensitive to the evaluator, we
additionally compute a Gemini-3.1-Pro-based counterpart of
Table~\ref{tab:benchmark_caption}. Under Gemini-3.1-Pro, \vlmname{} remains
the strongest model in both easy and hard settings, and the high-level
conclusion is unchanged even though absolute scores and some close baseline
orderings vary slightly.

\begin{table}[htbp]
  \caption{\textbf{Caption benchmark results on \benchmarkname{} (\%) under
  a Gemini-3.1-Pro judge.} We report the same easy/hard caption metrics as in
  the main paper, but use Gemini-3.1-Pro for caption-fact alignment.
  \textbf{Cons.}: Consistency; \textbf{Cov.}: Coverage; \textbf{A-Hal.}:
  Anti-Hallucination. Best value per column is \textbf{bold}.}
  \label{tab:benchmark_caption_gemini_judge}
  \centering
  \footnotesize
  \renewcommand{\arraystretch}{1.05}
  \begin{tabular*}{\linewidth}{@{\extracolsep{\fill}}lcccccccc@{}}
    \toprule
    \multirow{2}{*}{Model}
      & \multicolumn{4}{c}{Easy}
      & \multicolumn{4}{c}{Hard} \\
    \cmidrule(lr){2-5} \cmidrule(lr){6-9}
      & Overall$\uparrow$ & Cons.$\uparrow$ & Cov.$\uparrow$ & A-Hal.$\uparrow$ & Overall$\uparrow$ & Cons.$\uparrow$ & Cov.$\uparrow$ & A-Hal.$\uparrow$ \\
    \midrule
    Qwen3-VL-Plus       & 73.7 & 78.1 & 54.3 & 88.6 & 63.5 & 71.5 & 51.4 & 67.7 \\
    Qwen3.5-Plus        & 75.7 & 79.7 & 53.6 & 93.9 & 71.5 & 76.3 & 50.4 & 88.1 \\
    Doubao-Seed-2.0-Pro & 77.1 & 80.9 & 53.0 & \textbf{97.4} & 73.7 & 75.2 & 62.9 & 82.7 \\
    Gemini-3.1-Pro      & 78.1 & 80.0 & 63.7 & 90.6 & 73.7 & 78.5 & 53.6 & 88.8 \\
    GPT-5.4             & 80.5 & 81.4 & 71.2 & 88.7 & 78.0 & 77.4 & 66.6 & 90.2 \\
    \midrule
    \vlmname{} (Ours)   & \textbf{83.2} & \textbf{84.5} & \textbf{71.7} & 93.4 & \textbf{83.0} & \textbf{83.1} & \textbf{71.9} & \textbf{94.1} \\
    \bottomrule
  \end{tabular*}
\end{table}

\subsubsection{Human Alignment Study}
\label{sec:app_bench_human}
\label{app:human_ranking_protocol}

This subsection supports the human-alignment claim in
Section~\ref{sec:benchmark_results}. It explains how the 10-human-rater study
is constructed and how the benchmark scores are normalized before correlation
analysis.

We recruit 10 human raters. For each benchmark sample, annotators are shown the
robot video together with six candidate captions, one from each evaluated
model: Qwen3-VL-Plus, Qwen3.5-Plus, Doubao-Seed-2.0-Pro, Gemini-3.1-Pro,
GPT-5.4, and \vlmname{}. Each caption is ranked from 1 to 6, where 1 denotes
the best caption and 6 denotes the worst. The protocol is conducted on the
same 500 benchmark videos used for automatic caption evaluation.

Annotators jointly consider factual correctness, process coverage, temporal
coherence, object grounding, and resistance to hallucination. After annotation,
we average the assigned ranks across raters and samples to obtain a single
human score per model. These scores are normalized from the theoretical 1--6
range to $[0,1]$, while benchmark caption Overall scores are normalized from
0--100 to $[0,1]$. The resulting correlations reported in the main paper are
high in both settings: easy Pearson \textbf{0.937} and Spearman
$\rho$ \textbf{0.943}; hard Pearson \textbf{0.922} and Spearman
$\rho$ \textbf{0.943}.

\begin{figure}[t]
  \centering
  \begin{subfigure}[t]{0.49\linewidth}
    \centering
    \includegraphics[width=\linewidth]{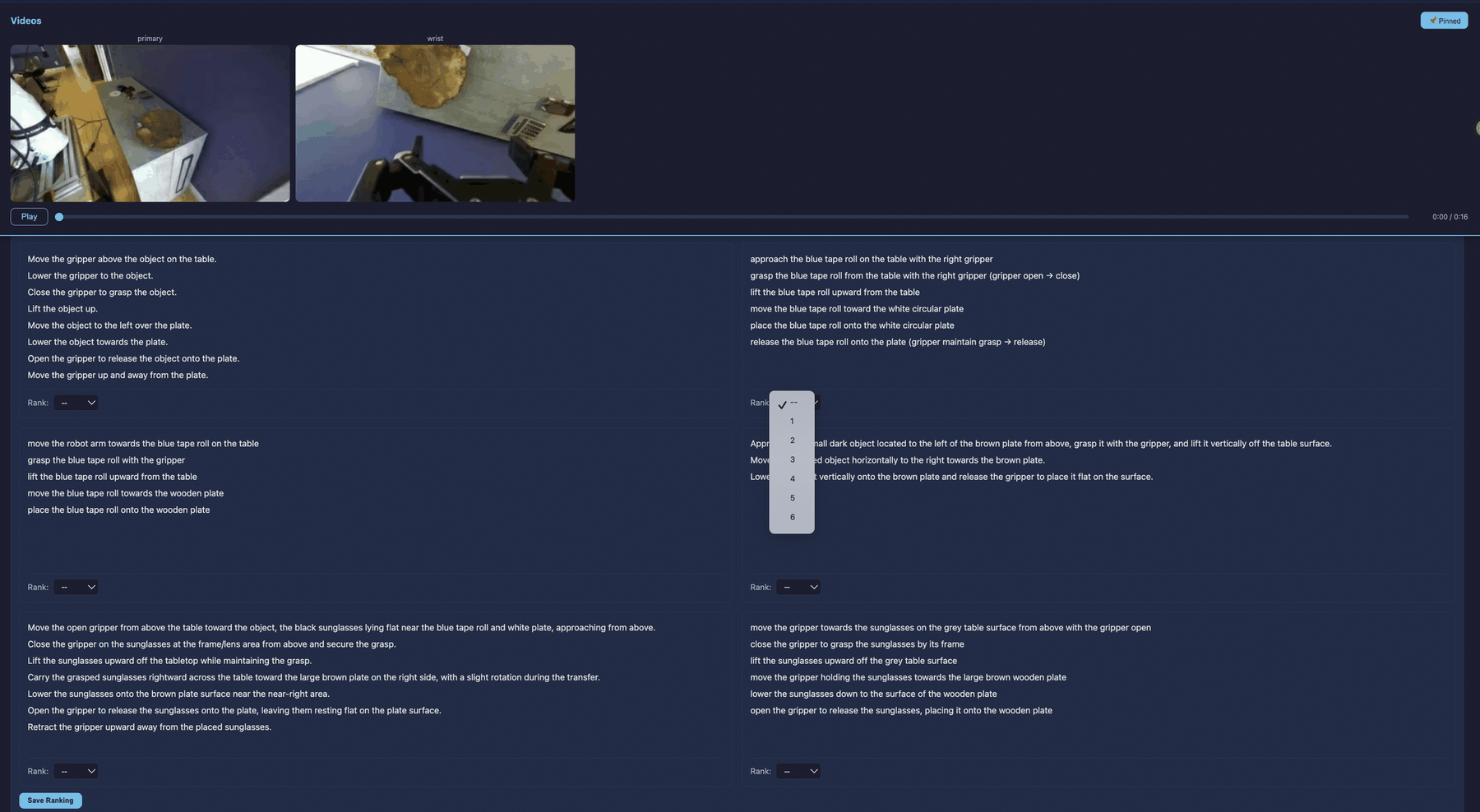}
    \caption{Human ranking interface on a short tabletop manipulation sample.}
  \end{subfigure}
  \hfill
  \begin{subfigure}[t]{0.49\linewidth}
    \centering
    \includegraphics[width=\linewidth]{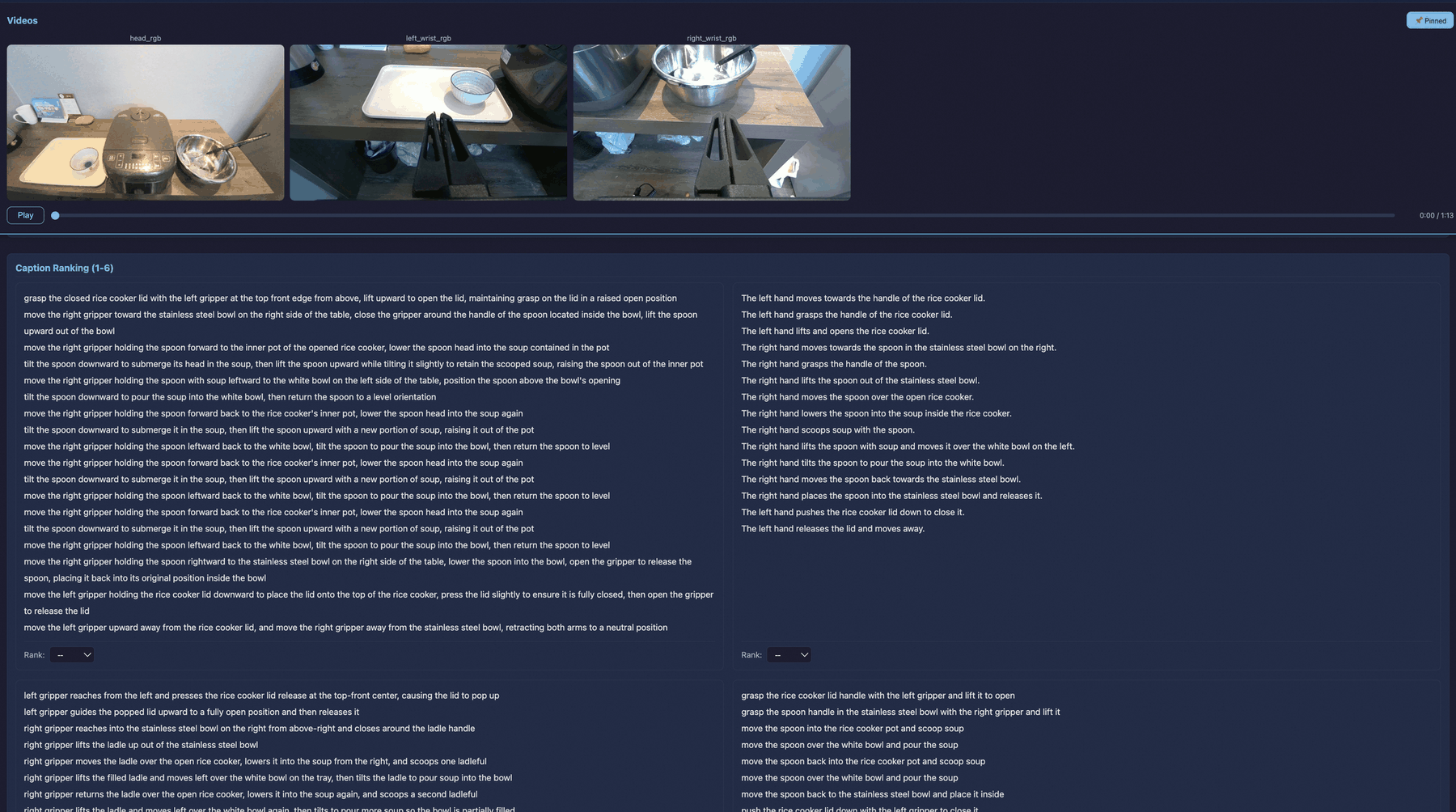}
    \caption{Human ranking interface on a longer multi-step kitchen task.}
  \end{subfigure}
  \caption{\textbf{Human ranking interface for caption evaluation.}
  Annotators watch the benchmark video and rank the six candidate captions from
  best to worst according to fine-grained faithfulness and usefulness. The
  protocol is designed to validate whether benchmark-induced model ranking is
  aligned with direct human judgment.}
  \label{fig:human_ranking_interface}
\end{figure}

\subsubsection{Caption Cost, Token, and Latency}
\label{sec:app_bench_cost}
\label{app:caption_cost}

This subsection supports the efficiency discussion in
Section~\ref{sec:benchmark_results}. It reports average token consumption and
elapsed time per sample for the caption track.

Table~\ref{tab:caption_token_cost} shows that \vlmname{} achieves the best
caption quality while remaining substantially more token-efficient than several
strong closed-source baselines, especially Doubao-Seed-2.0-Pro and
Gemini-3.1-Pro.

\begin{table}[htbp]
  \caption{\textbf{Caption-track inference cost.} Average total token
  consumption and average elapsed time per sample on the caption track.}
  \label{tab:caption_token_cost}
  \centering
  \small
  \begin{tabular}{@{} l cc @{}}
    \toprule
    \textbf{Model} & \textbf{Avg. total tokens} & \textbf{Avg. elapsed (s)} \\
    \midrule
    \vlmname{} & 14,438.1 & 74.5 \\
    GPT-5.4 & 21,906.6 & 62.2 \\
    Doubao-Seed-2.0-Pro & 93,792.2 & 104.8 \\
    Gemini-3.1-Pro & 198,271.9 & 62.8 \\
    Qwen3.5-Plus & 13,219.8 & 95.6 \\
    Qwen3-VL-Plus & 8,352.0 & 6.4 \\
    \bottomrule
  \end{tabular}
\end{table}

\subsubsection{Detailed Benchmark Results}
\label{sec:app_bench_analysis}
\label{app:benchmark_detailed_analysis}

This subsection supports the concise benchmark discussion in
Section~\ref{sec:benchmark_results}. It provides the more detailed result
interpretation moved out of the main paper to save space.

\paragraph{VQA analysis.}
\vlmname{} reaches \textbf{68.2\%} overall accuracy, outperforming the
strongest general-purpose baseline, GPT-5.4, by \textbf{8.0} absolute
points. The largest gain appears on Action and Motion Understanding, where
\vlmname{} improves from \textbf{64.6\%} to \textbf{75.7\%}, indicating a
substantial advantage in understanding execution order, contact patterns, and
motion dynamics. The gains are also consistent on Entity and Scene Grounding
(\textbf{64.7\%} $\rightarrow$ \textbf{72.5\%}) and Interaction and State
Reasoning (\textbf{57.1\%} $\rightarrow$ \textbf{68.5\%}), showing that the
benefit of fine-grained supervision extends beyond object recognition to
process-level reasoning.

Relative to its base model, Qwen3.5-Plus, supervised fine-tuning on
\datasetname{} raises overall VQA accuracy from \textbf{55.9\%} to
\textbf{68.2\%}. The improvement is broad-based across all three reporting
axes, with gains of \textbf{+8.6}, \textbf{+15.7}, and \textbf{+22.3} points
on Gnd., Act., and State, respectively. This comparison isolates the effect of
fine-grained action-aligned supervision more directly than cross-model
comparison alone, and shows that \datasetname{} substantially strengthens the
robotic video understanding capability of the underlying VLM.

\paragraph{Caption analysis.}
Instruction input is beneficial for all evaluated models: caption Overall is
consistently higher in the easy setting than in the hard setting. The easy-hard
gap is especially large for Qwen3-VL-Plus and Doubao-Seed-2.0-Pro, suggesting
that these models rely more heavily on the original task instruction. The same
trend appears in Anti-Hallucination: some general-purpose VLMs depend strongly
on instruction input to avoid fabricated action content, whereas \vlmname{}
remains comparatively stable when the instruction is removed.

\vlmname{} remains the strongest model overall. In the easy setting, it
achieves the best Overall, Consistency, and Coverage scores. In the hard
setting, which requires the model to infer the manipulation process directly
from video rather than from task-level language priors, \vlmname{} is SOTA on
all four metrics and improves Overall from the strongest baseline score of
\textbf{78.0\%} to \textbf{82.2\%}. These results indicate that the gain from
fine-grained supervision is not limited to instruction-conditioned captioning,
but extends to intrinsic process understanding.

\paragraph{Benchmark validity.}
Unless otherwise specified, the caption results in
Table~\ref{tab:benchmark_caption} use GPT-5.4-Pro as the alignment judge.
Re-evaluating the same captions with Gemini-3.1-Pro preserves \vlmname{} as
the strongest model in both easy and hard settings, despite small shifts in
absolute scores and close baseline orderings; the corresponding robustness
discussion is reported in
Appendix~\ref{app:caption_judge_robustness}. We further compare automatic
caption scores with human judgment by asking 10 raters to rank the six models
on the same 500 benchmark videos. As shown in
Figure~\ref{fig:caption_human_alignment}, the resulting agreement is strong in
both settings (easy: Pearson \textbf{0.937}, Spearman $\rho$ \textbf{0.943};
hard: Pearson \textbf{0.922}, Spearman $\rho$ \textbf{0.943}), indicating that
the caption track is both aligned with human preference and robust to the
choice of judge model.

\subsection{FineVLA-Policy Setup}
\label{sec:app_policy_setup}
\label{app:policy_setup}

This section provides implementation details for the policy setup used in
Section~\ref{sec:FineVLA}. It specifies the policy architectures, pretraining
configuration, RoboTwin fine-tuning data, and instruction-mixing design used
to study the effect of fine-grained language supervision.

\begin{table}[htbp]
\centering
\small
\caption{\textbf{Training configurations for \policyname{}.}
The table summarizes the backbone, training dataset, optimization length, and
compute budget used in the pretraining, RoboTwin fine-tuning, and real-world
fine-tuning stages. The variant names indicate the policy architecture.}
\label{tab:policy_training_configs}
\setlength{\tabcolsep}{3pt}
\renewcommand{\arraystretch}{1.05}
\begin{tabular*}{\linewidth}{@{\extracolsep{\fill}}p{1.05cm}p{2.1cm}p{1.65cm}p{2.1cm}ccc@{}}
\toprule
Stage & Variant & Backbone & Dataset & Steps & GPUs & Global BS \\
\midrule
Pretrain & RDT-OFT & Qwen3.5-4B & RDT & 100k & 64$\times$A100 & 512 \\
Pretrain & RDT-GR00T & Qwen3.5-4B & RDT & 100k & 64$\times$A100 & 512 \\
Pretrain & AlohaMix-OFT & Qwen3.5-4B & AlohaMix & 100k & 64$\times$A100 & 512 \\
\midrule
RoboTwin SFT & All pretrained variants & Qwen3.5-4B & RoboTwin & 100k & 8$\times$A100 & 128 \\
Real-World SFT & AlohaMix-OFT & Qwen3.5-4B & Real (600 demos) & 100k & 8$\times$A800 & 64 \\
\bottomrule
\end{tabular*}
\end{table}

\subsubsection{Policy Frameworks}
\label{sec:app_policy_frameworks}

We instantiate \policyname{} with two action-decoding frameworks implemented in
the StarVLA codebase. \textbf{StarVLA-GR00T} adopts a dual-system design where
the VL backbone serves as System~2 for slow reasoning and a DiT-based
flow-matching module serves as System~1 for fast action generation, consistent
with GR00T N1.5. \textbf{StarVLA-OFT} attaches a lightweight MLP head that
reads the hidden states of predefined action tokens and regresses continuous
actions in parallel with an L1 objective, following OpenVLA-OFT. The two
variants share the same Qwen3.5-4B backbone and the same visual observations
and language inputs.

\subsubsection{Pretraining Datasets and Configuration}
\label{sec:app_policy_datasets}

We first pretrain three policy variants for 100k steps: \textbf{RDT-OFT},
\textbf{RDT-GR00T}, and \textbf{AlohaMix-OFT}. Here, OFT and GR00T denote the
policy architecture, while RDT and AlohaMix denote the training dataset.
\textbf{AlohaMix} is an ALOHA-only mixture constructed from open-source datasets
such as RoboCOIN and RoboMIND, and contains 86,662 episodes across 598 tasks,
approximately 13$\times$ larger than RDT. All pretraining runs use 64 A100
GPUs for 100k steps, with per-device batch size 8 and global batch size 512.

Table~\ref{tab:alohamix_composition} details the composition of
AlohaMix. We deliberately restrict the mixture to ALOHA-compatible
dual-arm embodiments so that all trajectories share the same
kinematic structure. This single-embodiment design avoids
cross-embodiment confounds and lets us attribute performance
differences solely to the language supervision signal. Of the
86,662 total episodes, 5,872 have fine-grained annotations
produced by \toolname{} and verified by human annotators; these
form the FG dataset used in the instruction-mixing experiments
(Section~\ref{sec:instruction_mixing}).

\begin{table}[htbp]
  \centering
  \small
  \caption{\textbf{AlohaMix pretraining dataset composition.}
  All sources use ALOHA-compatible dual-arm embodiments. FG
  annotations denote the number of episodes with human-verified
  fine-grained instructions from \datasetname{}. Because AlohaMix
  retains only the ALOHA-compatible dual-arm subset of each source,
  these per-source FG counts are subsets of the corresponding
  full-source representative counts in
  Table~\ref{tab:dataset_stats}.}
  \label{tab:alohamix_composition}
  \begin{tabular}{@{} l r r r r @{}}
    \toprule
    Source & Tasks & Episodes & Frames & FG Ann. \\
    \midrule
    RDT               &  296 &   6,061 &   3,169,041 & 1,275 \\
    RoboCOIN          &   81 &  35,151 &  23,035,912 & 2,907 \\
    RoboMIND-V1.0     &   78 &   9,370 &   5,891,176 &   594 \\
    RoboMIND-V2.0     &  143 &  36,080 &  74,502,637 & 1,096 \\
    \midrule
    \textbf{Total}
      & \textbf{598} & \textbf{86,662}
      & \textbf{106,598,766} & \textbf{5,872} \\
    \bottomrule
  \end{tabular}
\end{table}

\subsubsection{RoboTwin Fine-Tuning Data and Configuration}
\label{sec:app_policy_robotwin_sft}

We fine-tune the pretrained policies on RoboTwin using the union of the Clean
and Random training sets. The resulting supervised fine-tuning corpus contains
27,500 trajectories and 6,075,103 transitions. All RoboTwin fine-tuning runs
use 8 A100 GPUs for 100k steps, with per-device batch size 16 and global batch
size 128.

\subsubsection{Instruction Mixing Construction}
\label{sec:app_instruction_mixing}

For each dataset and architecture, we keep the robot trajectories, action
labels, visual observations, and all other training signals fixed; the
\emph{only} variable is the language instruction paired with each trajectory.
Every trajectory has two instruction variants: a \textbf{Raw} goal-level
instruction (e.g., the original task name) and a \textbf{FG} fine-grained
process-level description generated by \toolname{}.

The FG:Raw ratio controls the \emph{sampling probability} during training.
For example, FG:Raw~$=$~1:4 means that each trajectory has a $\frac{1}{5}$
probability of being paired with its FG instruction and a $\frac{4}{5}$
probability of being paired with its Raw instruction when sampled for a
training step. The trajectory itself, its action labels, and visual
observations remain identical regardless of which instruction is drawn. This
ensures that observed performance differences are attributable solely to the
instruction type.

We compare seven configurations: Raw-only, FG:Raw~$=$~1:4, 1:2, 1:1, 2:1,
4:1, and FG-only. This design isolates the effect of action-aligned language
supervision from changes in data scale, embodiment, or action distribution.

\subsubsection{Additional Training Details}
\label{sec:app_policy_hparams}

We use the same backbone, observation interface, and action interface across
all instruction-mixing settings within each framework. Real-world training-set
construction and optimizer-level hyperparameters will be documented once the
final runs are locked.

\subsection{RoboTwin Details and Additional Analysis}
\label{sec:app_robotwin}
\label{app:robotwin_analysis}

This section supports the RoboTwin results in Section~\ref{sec:policy_robotwin}.
It provides the evaluation protocol, the full mixing-ratio result table, and
the compact analyses that explain the inverted-U trend and the interaction with
architecture and dataset scale.

\subsubsection{RoboTwin Evaluation Protocol}
\label{sec:app_robotwin_protocol}

This subsection supports the evaluation setup in Section~\ref{sec:policy_robotwin}.
We evaluate policies on the official RoboTwin \textbf{Easy} and \textbf{Hard}
splits and report success rate averaged over 20 episodes per task. A trial is
counted as successful only if the task-specific goal condition is completed at
the end of the rollout; reported scores are the average success rates on the
corresponding split.

\subsubsection{Full RoboTwin Results}
\label{sec:app_robotwin_full}

This subsection provides the complete mixing-ratio table underlying the
RoboTwin discussion in the main paper.

\begin{table}[htbp]
  \caption{\textbf{Full RoboTwin simulation success rates (\%).} We compare
  three training settings (RDT-OFT, RDT-GR00T, and AlohaMix-OFT) under seven
  FG:Raw instruction ratios. Easy/Hard follow the official RoboTwin splits.
  Best value per column is \textbf{bold}.}
  \label{tab:app_robotwin_full}
  \centering
  \small
  \renewcommand{\arraystretch}{1.05}
  \begin{tabular*}{\linewidth}{@{\extracolsep{\fill}}lcccccc@{}}
    \toprule
    \multirow{2}{*}{\textbf{FG:Raw}}
      & \multicolumn{2}{c}{\textbf{RDT-OFT}}
      & \multicolumn{2}{c}{\textbf{RDT-GR00T}}
      & \multicolumn{2}{c}{\textbf{AlohaMix-OFT}} \\
    \cmidrule(lr){2-3} \cmidrule(lr){4-5} \cmidrule(lr){6-7}
      & Easy$\uparrow$ & Hard$\uparrow$ & Easy$\uparrow$ & Hard$\uparrow$ & Easy$\uparrow$ & Hard$\uparrow$ \\
    \midrule
    Raw-only                    & 61.5 & 60.0 & 55.1 & 53.4 & 71.8 & 71.4 \\
    FG\,:\,Raw $=$ 1\,:\,4    & 68.2 & 66.5 & 58.2 & 55.7 & 75.3 & 74.3 \\
    FG\,:\,Raw $=$ 1\,:\,2    & \textbf{74.1} & 72.1 & 61.7 & 60.9 & 82.8 & 78.6 \\
    FG\,:\,Raw $=$ 1\,:\,1    & 73.9 & \textbf{72.4} & \textbf{69.4} & \textbf{68.2} & \textbf{86.8} & \textbf{82.5} \\
    FG\,:\,Raw $=$ 2\,:\,1    & 70.4 & 68.3 & 65.9 & 63.1 & 80.9 & 79.3 \\
    FG\,:\,Raw $=$ 4\,:\,1    & 68.6 & 67.5 & 64.9 & 63.2 & 79.5 & 78.5 \\
    FG-only                     & 62.9 & 62.0 & 62.1 & 61.5 & 78.3 & 76.1 \\
    \bottomrule
  \end{tabular*}
\end{table}

\subsubsection{Mixing-Ratio Analysis}
\label{sec:app_robotwin_mixing}

This subsection supports the mixing-ratio interpretation in
Section~\ref{sec:policy_robotwin}. Across all three evaluated settings,
success rate first rises and then falls as the FG proportion increases, with a
peak around FG\,:\,Raw $=$ 1\,:\,2 to 1\,:\,1. The trend is therefore
consistent with an inverted-U relationship rather than a monotonic preference
for either raw-only or FG-only supervision. Empirically, this means that raw
instructions primarily specify \emph{what} to achieve, while fine-grained
instructions specify \emph{how} to achieve it; the best performance emerges
when both signals are present.

\subsubsection{Architecture and Scale Analysis}
\label{sec:app_robotwin_arch_scale}

This subsection provides additional analysis on architectural sensitivity and
data-scale interaction, complementing the main findings in
Section~\ref{sec:policy_robotwin}.

Table~\ref{tab:robotwin_analysis_summary} summarizes the detailed numbers used
in the main-paper discussion. Panel~A reports the three representative
supervision regimes for each (dataset, framework) setting: Raw-only, the best
mixed FG:Raw ratio, and FG-only. Panel~B reports the derived comparison gaps
used to support the architectural-equalization and data-scale analyses.

\begin{table*}[htbp]
  \caption{\textbf{RoboTwin analysis summary.} Panel~A lists the raw-only,
  best mixed, and FG-only performance for each setting. Panel~B reports the
  derived gaps used in the architectural and data-scale comparisons. All
  values are success rates (\%).}
  \label{tab:robotwin_analysis_summary}
  \centering
  \small
  \setlength{\tabcolsep}{4pt}
  \resizebox{\linewidth}{!}{%
  \begin{tabular}{@{}lcccccc@{}}
    \toprule
    \multicolumn{7}{@{}l}{\textbf{Panel A: Representative supervision regimes}} \\
    \midrule
    \textbf{Setting} & \textbf{Raw-only} & \textbf{Best mixed ratio}
      & \textbf{Best mixed} & \textbf{FG-only}
      & \textbf{$\Delta$ (Raw$\rightarrow$Best)} & \textbf{$\Delta$ (Raw$\rightarrow$FG)} \\
    \midrule
    RDT-OFT
      & 61.5 / 60.0
      & 1:2 / 1:1
      & 74.1 / 72.4
      & 62.9 / 62.0
      & +12.6 / +12.4
      & +1.4 / +2.0 \\
    RDT-GR00T
      & 55.1 / 53.4
      & 1:1 / 1:1
      & 69.4 / 68.2
      & 62.1 / 61.5
      & +14.3 / +14.8
      & +7.0 / +8.1 \\
    AlohaMix-OFT
      & 71.8 / 71.4
      & 1:1 / 1:1
      & 86.8 / 82.5
      & 78.3 / 76.1
      & +15.0 / +11.1
      & +6.5 / +4.7 \\
    \midrule
    \multicolumn{7}{@{}l}{\footnotesize Values are reported as Easy / Hard.} \\
    \midrule
    \multicolumn{7}{@{}l}{\textbf{Panel B: Derived comparison gaps}} \\
    \midrule
    \textbf{Comparison} & \textbf{Raw-only gap} & \textbf{Best mixed gap}
      & \textbf{FG-only gap} & \multicolumn{3}{l@{}}{\textbf{Interpretation}} \\
    \midrule
    OFT -- GR00T on RDT
      & 6.4 / 6.6
      & 4.7 / 4.2
      & 0.8 / 0.5
      & \multicolumn{3}{l@{}}{Framework gap narrows as FG ratio increases.} \\
    AlohaMix -- RDT under OFT
      & 10.3 / 11.4
      & 12.7 / 10.1
      & 15.4 / 14.1
      & \multicolumn{3}{l@{}}{FG benefit is larger at bigger data scale.} \\
    \bottomrule
  \end{tabular}
  }
\end{table*}

\subsection{Real-World Policy Details}
\label{sec:app_realworld}
\label{app:policy_realworld}

This section supports the real-world policy experiments in
Section~\ref{sec:policy_real}. It records the task definitions and protocol
used for the real dual-arm evaluation; the quantitative results are reported
in the main text (Table~\ref{tab:real}).

\subsubsection{Robot Hardware and Setup}
\label{sec:app_real_setup}

This subsection supports the real-world evaluation setting in
Section~\ref{sec:policy_real}.

\paragraph{Hardware.}
We use a Cobot Magic dual-arm robot with three synchronized RGB cameras
(two wrist-mounted, one third-person). The action space consists of 14
joint-position commands (7 per arm) and two continuous gripper commands.
Policies are trained with action chunks of length 50 and executed
asynchronously at inference time. The low-level controller runs at
30\,Hz, and inference is served remotely on an 8$\times$A800 GPU server.

\paragraph{Training data.}
We collect 50 teleoperated demonstrations for each of 12 tabletop tasks
(600 episodes total). All demonstrations are recorded with joint-space
actions at 30\,Hz and synchronized multi-view video. We train a single
language-conditioned policy starting from the pretrained checkpoint
(Section~\ref{sec:FineVLA}) and fine-tune for 100k steps on
8$\times$A800 (80\,GB) GPUs with global batch size 64
($\sim$1.5 days wall-clock time).

\paragraph{Inference.}
At test time, the policy receives three RGB frames (one per camera) and
a language instruction, and outputs an action chunk of 50 steps. Actions
are dispatched asynchronously to the low-level controller. No
post-processing or action filtering is applied.

\subsubsection{Real-World Tasks}
\label{sec:app_real_tasks}

This subsection supports the task selection described in
Section~\ref{sec:policy_real}. The real-world evaluation suite
contains two general manipulation tasks, five in-distribution
instruction-sensitive task families (each comprising a paired
variant), and one out-of-distribution compositional probe, each
probing a specific control factor:
\textit{Clean Table} and \textit{Stack Block} (routine
manipulation),
\textit{Color} (object color grounding),
\textit{Pose} (initial-state grounding),
\textit{Approach} (approach direction),
\textit{Rotate} (rotation direction),
\textit{Arm} (active-arm selection, R\,$\to$\,R / L\,$\to$\,L),
and \textit{Arm+Target} (active-arm selection with unseen
actor-target binding, OOD probe).
Table~\ref{tab:real_tasks} lists the paired variants and their
corresponding language instructions.

\input{tables/real_tasks}

\subsubsection{Real-World Evaluation Protocol}
\label{sec:app_real_protocol}

This subsection supports the real-world partial-score numbers reported in the
main paper. Each task is evaluated over 10 trials. A trial is scored by
manually checking ordered subgoals; a completed subgoal receives proportional
credit. Between trials, the scene is reset to the designated initial
configuration before the next evaluation begins.

\subsubsection{Subgoal Definitions for Partial Scoring}
\label{app:real_subgoals}

Each task is decomposed into ordered subgoals. A trial receives credit
proportional to the fraction of completed subgoals.
Table~\ref{tab:subgoal_defs} lists the subgoal sequence for each task.
The \textbf{language-critical subgoal} (marked with $\star$) is the one
whose completion requires resolving the fine-grained instruction factor.

\begin{table}[h]
  \caption{\textbf{Subgoal definitions for real-world partial scoring.}
    $\star$ marks the language-critical subgoal for each task.}
  \label{tab:subgoal_defs}
  \centering
  \small
  \begin{tabular}{@{}lp{9.5cm}@{}}
    \toprule
    \textbf{Task} & \textbf{Ordered Subgoals} \\
    \midrule
    Clean Table &
      Reach $\to$ Grasp $\to$ Lift $\to$ Place in bin \\
    Stack Block &
      Identify target block $\to$ Reach $\to$ Grasp $\to$ Transport $\to$ Align \& Stack \\
    Color &
      $\star$Identify correct color $\to$ Reach $\to$ Grasp $\to$ Lift $\to$ Place in pen cup \\
    Pose &
      $\star$Identify pose $\to$ Approach from correct angle $\to$ Grasp $\to$ Lift $\to$ Place in box \\
    Approach &
      $\star$Approach from the specified direction $\to$ Grasp $\to$ Lift $\to$ Move over pink bowl $\to$ Release \\
    Rotate &
      Grasp $\to$ $\star$Rotate 90\textdegree{} in the specified direction $\to$ Release \\
    Arm &
      $\star$Select the correct arm $\to$ Reach $\to$ Grasp $\to$ Transport $\to$ Place in the correct bowl \\
    Arm+Target (OOD) &
      $\star$Select left arm $\to$ Reach $\to$ Grasp $\to$ Transport $\to$ Place in right bowl \\
    \bottomrule
  \end{tabular}
\end{table}

\subsection{Additional Analysis}
\label{sec:app_ablation}

This section collects supplementary observations from the RoboTwin experiments
that are useful for interpretation but not required for the main narrative.

\paragraph{FG supervision narrows the architecture gap.}
Comparing StarVLA-OFT and StarVLA-GR00T on the same dataset (RDT), OFT is
clearly stronger under Raw-only supervision (gap of 6.4/6.6 on Easy/Hard), but
the gap shrinks as FG ratio increases and nearly vanishes under FG-only
(0.8/0.5). This suggests that dense language supervision alleviates a
supervision bottleneck, reducing the policy's dependence on decoder
architecture choice.

\paragraph{FG supervision benefits more from larger data scale.}
Comparing RDT-OFT and AlohaMix-OFT, the gain from FG supervision is larger
on the bigger AlohaMix dataset. The FG-only improvement over Raw-only grows
from +1.4/+2.0 (RDT) to +6.5/+4.7 (AlohaMix). As trajectory diversity grows,
dense action-aligned language has more distinct patterns to bind to, suggesting
that FG supervision should become even more valuable at larger training scale.

Detailed per-setting numbers supporting both observations are reported in
Table~\ref{tab:robotwin_analysis_summary}.

\subsection{Reproducibility, Limitations, and Ethics}
\label{sec:app_limitations}
\label{app:limitations}

This section supports the reproducibility, limitation, and ethics-related
claims referenced by the NeurIPS checklist. It summarizes the compute budget
used in the main training experiments, clarifies the current limitations of the
pipeline, and records the intended release scope.

\subsubsection{Reproducibility Checklist Support}
\label{sec:app_repro_support}

The main paper and appendix together document the data-construction pipeline,
benchmark protocol, policy architectures, instruction-mixing setup, and the
training configurations used in the experiments. In particular,
Appendix~\ref{app:tool} details the construction of \toolname{},
Appendix~\ref{app:benchmark} details \benchmarkname{}, and
Appendix~\ref{app:policy_setup} details the policy-training setup used in
RoboTwin and the real-world experiments.

\subsubsection{Compute Resources}
\label{sec:app_compute}

\vlmname{} supervised fine-tuning is performed for 903 steps on 256 NVIDIA
H200 GPUs with global batch size 512, learning rate decayed from 7e-6 to 7e-7,
taking approximately 40 hours and using roughly 105\,GB of memory per GPU.
Policy pretraining is performed for 100k steps on 64 A100 GPUs with global
batch size 512, taking approximately 48 hours per run and using roughly
70\,GB of memory per GPU. RoboTwin fine-tuning is performed for 100k steps on
8 A100 GPUs with global batch size 128, taking approximately 48 hours per run
and using roughly 75\,GB of memory per GPU. Real-world fine-tuning is performed
for 100k steps on 8$\times$A800 (80\,GB) GPUs with global batch size 64,
taking approximately 1.5 days per run.

\subsubsection{Limitations}
\label{sec:app_limitation_notes}

The current work has two main limitations. First, although \vlmname{} provides
high-quality scalable annotation, a small portion of generated annotations
still requires manual verification, so the pipeline does not fully eliminate
human-in-the-loop supervision. Second, while the method is validated on
multiple datasets, RoboTwin, and a set of real-world tasks, broader validation
across additional robot embodiments and a larger set of real-world tasks
remains future work.

\subsubsection{Societal Impact and Safety}
\label{sec:app_societal}

Fine-grained language supervision can improve the controllability and
transparency of robot behavior by making execution constraints more explicit.
At the same time, deployment on real robotic systems still requires external
safety constraints, because incorrect grounding, hallucinated action details,
or control failures may lead to unintended physical interactions.

%% file: tables/real_tasks.tex
\begin{table*}[t]
  \caption{\textbf{Real-world evaluation tasks and paired variants.}
    Each instruction-sensitive factor is tested with two
    complementary language variants under the same visual scene.
    $^\dagger$\,L\,$\to$\,R uses an unseen actor-target binding
    (OOD probe).}
  \label{tab:real_tasks}
  \centering
  \small
  \setlength{\tabcolsep}{3.5pt}
  \footnotesize
  \begin{tabular}{@{}l p{5.2cm} p{5.2cm} c@{}}
    \toprule
    \textbf{Factor}
      & \textbf{Instruction A}
      & \textbf{Instruction B}
      & \textbf{Type} \\
    \midrule
    Clean Table
      & Clean up the table.
      & ---
      & General \\
    Stack Block
      & Stack the blue block on top of the red block.
      & ---
      & General \\
    \midrule
    Color
      & Put the red pen into the pen cup.
      & Put the blue pen into the pen cup.
      & ID \\
    Pose
      & Pick up the pen lying on the table and place it into the box.
      & Pick up the pen standing on the table and place it into the box.
      & ID \\
    Approach
      & Grasp the block from above, move it over the pink bowl, and release it.
      & Grasp the block from the right side, move it over the pink bowl, and release it.
      & ID \\
    Rotate
      & Rotate the pen clockwise for 90 degrees.
      & Rotate the pen counter-clockwise for 90 degrees.
      & ID \\
    Arm
      & Right hand pick up the block and place it into the right bowl.
      & Left hand pick up the block and place it into the left bowl.
      & ID \\
    \midrule
    Arm+Target$^\dagger$
      & Right hand pick up the block and place it into the right bowl.
      & Left hand pick up the block and place it into the right bowl.
      & OOD \\
    \bottomrule
  \end{tabular}
\end{table*}